\documentclass{article}

\usepackage[nonatbib, preprint]{neurips_2022}

\usepackage[utf8]{inputenc} 
\usepackage[T1]{fontenc}    
\usepackage{url}            
\usepackage{booktabs}       
\usepackage{amsfonts}       
\usepackage{nicefrac}       
\usepackage{microtype}      
\usepackage{xcolor}         
\usepackage{microtype}
\usepackage{graphicx}
\usepackage{subfigure}
\usepackage{booktabs} 
\usepackage{bbm}
\usepackage{mathtools}
\usepackage{algorithmic}
\usepackage{algorithm} 
\usepackage{amsmath}
\usepackage{amssymb}
\usepackage{mathtools}
\usepackage{amsthm}
\usepackage{multirow}
\usepackage{array}
\usepackage{makecell}
\usepackage{lscape}
\usepackage{pbox}

\title{Diffusion Transport Alignment}

\author{%
 Andres F. Duque\\
  Department of Mathematics and Statistics\\
  Utah State University\\
  \texttt{andres.duque@usu.edu} \\
  \And
   Guy Wolf\\
   Department of Mathematics and Statistics\\
 Université de Montréal\\
   \texttt{wolfguy@mila.quebec} \\
  \AND
  Kevin R. Moon \\
  Department of Mathematics and Statistics\\
  Utah State University\\ 
  \texttt{kevin.moon@usu.edu} \\
}

\begin{document}

\maketitle

\begin{abstract}
The integration of multimodal data presents a challenge in cases when the study of a given phenomena by different instruments or conditions generates distinct but related domains. Many existing data integration methods assume a known one-to-one correspondence between domains of the entire dataset, which may be unrealistic. Furthermore, existing manifold alignment methods are not suited for cases where the data contains domain-specific regions, i.e., there is not a counterpart for a certain portion of the data in the other domain. We propose Diffusion Transport Alignment (DTA), a semi-supervised manifold alignment method that exploits prior correspondence knowledge between only a few points to align the domains. By building a diffusion process, DTA finds a transportation plan between data measured from two heterogeneous domains with different feature spaces, which by assumption, share a similar geometrical structure coming from the same underlying data generating process. DTA can also compute a partial alignment in a data-driven fashion, resulting in accurate alignments when some data are measured in only one domain. We empirically demonstrate that DTA outperforms other methods in aligning multimodal data in this semisupervised setting. We also empirically show that the alignment obtained by DTA can improve the performance of machine learning tasks, such as domain adaptation, inter-domain feature mapping, and exploratory data analysis, while outperforming competing methods.
\end{abstract}

\section{Introduction}

In many data science applications, data may be collected from different measurement instruments, conditions, or protocols of the same underlying system. Examples include single cell RNA-sequencing and ATAC-sequencing measurements of the same group of cells \cite{stuart2019comprehensive}, text documents translated into different languages \cite{liu2016manifold}, brain images from multiple neuroimaging techniques \cite{vieira2020multimodal}, and images of an object or scene captured from different views \cite{hu2019mima}. 
In such settings, researchers are often interested in integrating data from the different domains to enhance our understanding of the system as well as the relationships between the different domains. Integrating the data may also lead to improved downstream analysis, such as classification, if there is domain-specific information about the task.

 Multi-view data integration is usually  performed assuming knowledge of one-to-one correspondences, i.e., the data comes in a paired fashion between domains. One of the simplest methods for this setting is Canonical Correlation Analysis (CCA), a linear approach that finds a projection that maximizes the correlation between the two domains~\cite{thompson1984canonical}. Kernel CCA extends this to nonlinear projections via the kernel trick~\cite{gao2012multi,chang2013canonical}. Alternating diffusion~\cite{katz2019alternating} and integrated diffusion~\cite{kuchroo2021multimodal} are nonlinear alignment methods  based on the robust manifold learning algorithm Diffusion Maps~\cite{coifman2006diffusion}. For an overview of other approaches see~\cite{gravina2017multi,lahat2015multimodal}.

A popular way to integrate distinct domains is manifold alignment. 
First introduced in the seminal works \cite{ham2003learning} and \cite{ham2005semisupervised}, this family of methods seeks to find projections of the multiple domains into a common latent space where inter-domain relationships can be captured. Manifold alignment can be performed in various scenarios, depending on how much information is provided about the correspondence between different domains. The edge case, usually referred to as \textit{unsupervised manifold alignment}, arises in the absence of any relationship known a priori between the domains as in  \cite{wang2009manifold}, \cite{cui2014generalized}, \cite{stanley2020harmonic}, \cite{cao2020unsupervised}, \cite{cao2022manifold} and \cite{demetci2022scot}. Some of the data integration approaches described previously, such as CCA, may be viewed as belonging to the opposite edge case of \textit{supervised manifold alignment}.

In contrast, there is a broad group of problems that can be categorized as  \textit{semi-supervised manifold alignment}. In this scenario, some degree of correspondence between domains is assumed to be known. In some cases, a one-to-one correspondence is known for only a few of the data points. This is the case in \cite{ham2005semisupervised}, which uses the Laplacian eigenmaps loss function in both domains while penalizing mismatches of known correspondences in the embedding. In \cite{wang2008manifold}, the authors first learn a latent representation for each domain using a variation of Laplacian eigenmaps \cite{belkin2003laplacian}. They then use Procrustes analysis in the common embedding space to find a transformation that aligns the matching observations, which subsequently is applied to the rest of the data.  This provides a final joint representation that accounts for the particular geometry of each domain, as well as the correspondence knowledge. Similarly, the approach proposed in \cite{lafon2006data} finds a low dimensional embedding generated by diffusion maps \cite{coifman2006diffusion} and then performs an affine transformation to align the known correspondences. More recently, a generative adversarial network  called manifold alignment GAN (MAGAN) was introduced in \cite{amodio2018magan}. MAGAN is based on a similar architecture as cycleGAN \cite{zhu2017unpaired}, which learns functions that map from one domain to another. However, the authors of MAGAN showed that cycleGAN and similar approaches tend to superimpose rather than align the data manifolds, resulting in incorrect alignments between distinct groups.  To mitigate this issue, MAGAN incorporates a correspondence loss between the known correspondences enforcing a consistent alignment.

Alternatively, the correspondence information may be available at the feature level. MAGAN can be applied to this case with a correspondence loss imposed on the shared features. Other approaches use class labels in both domains as the correspondence knowledge, as in \cite{wang2011heterogeneous} where the labels act as anchors points for the alignment. This was further expanded to a kernelized version in \cite{tuia2016kernel}.

 



In this work we focus on the semi-supervised  problem where we assume a known one-to-one correspondence between domains is available for a few of the data points. Our method, called Diffusion Transport Alignment (DTA), starts by building a diffusion process  \cite{coifman2006diffusion} that connects measurements in different domains via the known correspondences. In this fashion, DTA transforms both domains to a shared embedding space, allowing us to extract inter-domain distances. Finally, DTA solves a partial optimal transport problem which finds a coupling between data samples from one domain to the other. The newly obtained coupling can be further used to improve the performance of downstream analysis. For instance, one may be interested in learning a mapping between both domains, but the known correspondences are insufficient to successfully train a model. Another use-case is to perform unsupervised multi-domain analysis with methods as in \cite{lindenbaum2020multi} or \cite{katz2019alternating}, which require one-to-one correspondences between all points in all domains. DTA is also useful for domain adaptation, where a model is trained on a source domain and then applied to a target domain. 

Our contributions are as follows: 1) We develop a manifold alignment method, DTA, that outperforms current methods in recovering inter-domain relationships. 2) DTA can perform a data-driven partial alignment when a subset of the data is domain-specific, preventing spurious couplings between domains. 3) DTA can also include available label information to improve the performance, whereas competing methods do not. 4) We demonstrate the use of DTA in multiple applications.

\section{Diffusion Transport Alignment}
    

Consider a multi-domain data collection of a data generating process where two different views in potentially different feature spaces $\Phi_{1} \in \mathrm{R}^{n \times q}$ and $\Phi_{2} \in \mathrm{R}^{m \times p}$ are measured, containing observations $\{x_{i}\}_{i=1}^n$, and $\{y_{i}\}_{i=1}^m$, respectively.  We wish to learn a correspondence between both domains in a semi-supervised setting, where one-to-one correspondence is known for a set of observations denoted by $\mathcal{C}$. That is, for each $c \in \mathcal{C}$ we have access to its features in both domains. 

\begin{figure}[!h]
\begin{center}
\centerline{\includegraphics[width=\textwidth]{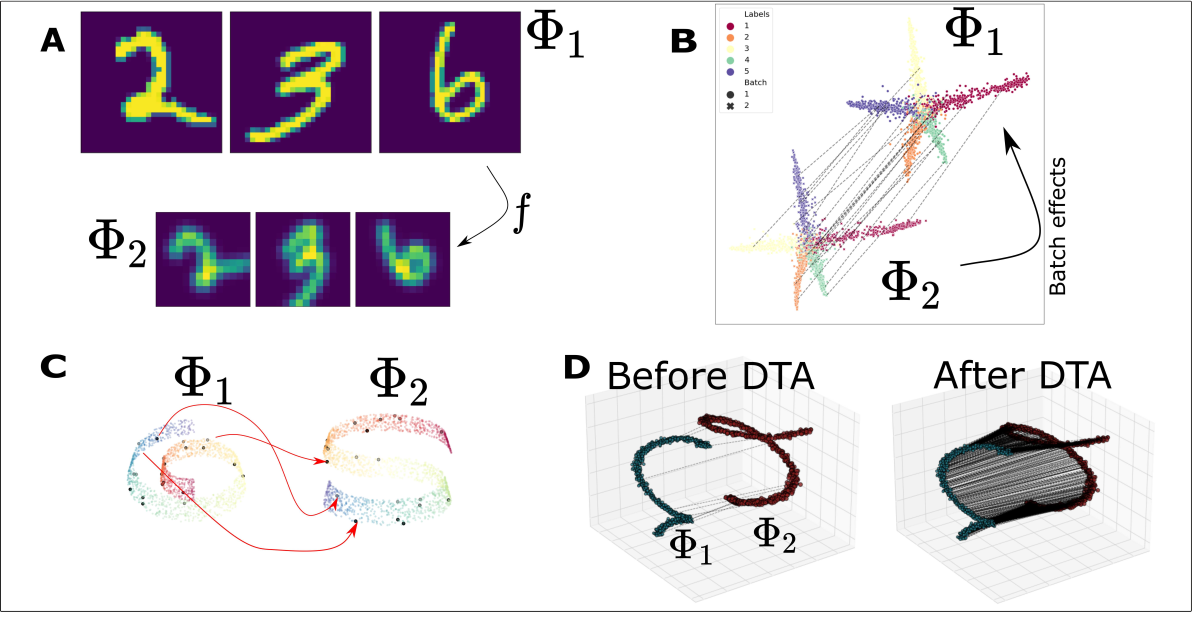}} 
\caption{\textbf{Motivating examples for DTA}. In all of these examples we have data measured in two distinct domains $\Phi_{1}$ and $\Phi_{2}$, and we possess a small subset of matching observations $\mathcal{C}$. This corresponds to the scenario where obtaining corresponding measurements may be costly, e.g. via expert annotation. The goal of DTA is to leverage the small subset of known correspondences to align the remaining observations. \textbf{A) Distorted MNIST digits}. Here $\Phi_{1}$ consists of the original MNIST digits, while $\Phi_{2}$ consists of distorted images after applying multiple transformations: rotation, downscaling, and Gaussian blurring. To learn a parametric function that maps from one domain to the other, the small set of correspondences is not enough. Thus, we need to find a greater set of matching data.
\textbf{B) Splatter simulation with batch effects~\cite{zappia2017splatter}}. A common problem when dealing with biological data is the distortion produced by the measurement protocols, introducing what is known as batch effects. Accurate alignment would overcome theses batch effects.  \textbf{C) Swiss roll and S curve}. This case presents the ideal scenario where the two domains are a smooth mapping from a common latent space. Black points indicate correspondences with  three of them (red arrows) highlighted. \textbf{D) Two helixes}. Here we use a dataset from \cite{tuia2016kernel} and display the effect of DTA after leveraging the known correspondences to align both manifolds.} 
\label{fig:half_mnist}
\end{center}
\vskip -0.2in
\end{figure}

As a motivating example, consider a classification problem where both domains contain labeled data points for some shared classes (see Section~\ref{sub:Feature}). The two domains may contain distinct information that is relevant for classification. An example of this is in single cell data with both RNA-sequencing and ATAC-sequencing measurements. In this case, training on the aligned data will lead to improved performance compared with training on the domains separately. As another example, researchers may be interested in the relationships between variables measured in separate domains. Aligning the domains enables a larger dataset to obtain more accurate estimates of relationship measures such as the correlation coefficient or mutual information (see Section~\ref{sub:MI}).
Figure \ref{fig:half_mnist} presents a schematic description of the problem, and other use cases with synthetic data where DTA can be beneficial such as batch effects correction in biological data.

The fundamental idea of DTA consists of learning a diffusion process between domains through a small set of known correspondences. These correspondences form a link for inter-domain transitions. Ultimately, this diffusion process can be used to extract an inter-domain distance measure, providing a dissimilarity among the observations in both domains.  To do this, DTA first constructs a diffusion operator over each domain, denoted as $P_{\Phi_{1}}$ and $P_{\Phi_{2}}$. A standard approach is to first compute an affinity matrix using a kernel. We use the $\alpha$-decay kernel~\cite{moon2019visualizing}:
\begin{equation}
    \begin{split}
    K_{k,\alpha}(x_i,x_j)=\frac{1}{2}\exp&\left( -\frac{||x_i-x_j||^\alpha}{\sigma_{k}^\alpha(x_i)}\right) + \frac{1}{2}\exp\left( -\frac{||x_i-x_j||^\alpha}{\sigma_{k}^\alpha(x_j)}\right), 
    \label{eq:DecayGaussKern}
    \end{split}
\end{equation}
where $\sigma_k(x_i)$ is the $k$-nearest neighbor distance of $x_i$ and $\alpha>0$. This kernel has two hyper-parameters $\alpha$ and $k$, which provide a trade-off between connectivity in the graph and local geometry preservation. Methods that employ this kernel are typically robust to the choice of these hyper-parameters~\cite{moon2019visualizing}.
The diffusion operator $P$ is then computed by row-normalizing the kernel matrix. In this way $P$ can be viewed as a probability transition matrix, representing a  Markov chain between observations. The probabilities of transitioning from one point to any other within a $t$ step random walk are obtained by powering the diffusion operator $P^{t}$.

Transitioning via the diffusion process between observations in separate domains requires each point to be within a $t$-step random walk of at least one observation in $\mathcal{C}$. DTA computes the transition probabilities between observations in $\Phi_{1}$ and $\Phi_{2}$ and elements in $\mathcal{C}$ in their respective domain by diffusing the process several steps, obtaining $P_{\Phi_{1}}^{t}$ and $P_{\Phi_{2}}^{t}$. The entries $(i,c)$ of  $P_{\Phi_{k}}^{t}$ with $c \in \mathcal{C}$ contain  the  transition probabilities from each observation   $i \in \Phi_{k}$ to the observations in $\mathcal{C}$. Thus, we can extract the columns and rows of $P_{\Phi_{1}}^{t}$ and $P_{\Phi_{2}}^{t}$ associated with the elements in $\mathcal{C}$, obtaining the submatrices: $\Gamma_{\Phi_{1}} \in \mathrm{R}^{n \times |\mathcal{C}|}$, $\Gamma_{\Phi_{2}} \in \mathrm{R}^{m \times |\mathcal{C}|}$, $\tilde{\Gamma}_{\Phi_{1}} \in \mathrm{R}^{ |\mathcal{C}| \times n}$, $\tilde{\Gamma}_{\Phi_{2}} \in \mathrm{R}^{|\mathcal{C}| \times m}$. 


Diffusion between domains is then connected by $P_{\Phi_{1}\Phi_{2}} := \Gamma_{\Phi_{1}}\tilde{\Gamma}_{\Phi_{2}}$. In this way, its  row-normalized version $\tilde{P}_{\Phi_{1}\Phi_{2}}$ contains the transition probabilities of each pair of observation from $\Phi_{1}$ to $\Phi_{2}$ via the points in $\mathcal{C}$. Analogously, $\tilde{P}_{\Phi_{2}\Phi_{1}}$ is constructed to provide the transitions in the opposite direction. This construction provides a natural way to compute inter-domain distances via a cosine distance:  
\begin{equation}
    \begin{split}
    D_{ij} = \left( 1 - \frac{\langle \tilde{P}_{\Phi_{1}\Phi_{2}}(i,:) , P^{t}_{\Phi_{2}}(j,:)\rangle}{||\tilde{P}_{\Phi_{1}\Phi_{2}}(i,:)||||P^{t}_{\Phi_{2}}(j,:)||}\right) + \left(1 - \frac{\langle \tilde{P}_{\Phi_{2}\Phi_{1}}(i,:) , P^{t}_{\Phi_{1}}(j,:)\rangle}{||\tilde{P}_{\Phi_{2}\Phi_{1}}(i,:)||||P_{\Phi_{1}}(j,:)||}\right),
    \label{eq:interDistances}
    \end{split}
\end{equation}
where, for example,  $P_{\Phi_{1}\Phi_{2}}(i,:)$ indicates the $i$th row in the matrix $P_{\Phi_{1}\Phi_{2}}$. We resort to cosine over euclidean distances between the diffusion operators since it resulted in a superior performance. 

The matrix $D$ contains inter-domain distances, but does not provide a direct alignment of the domains. The final step in DTA is to solve a partial optimal transport problem with $D$ as the cost matrix: 
\begin{align}
\begin{split}
    \min_T \quad & \sum_{i=1}^n\sum_{j=1}^m D_{ij}T_{ij}\\
    \text{s.t.} \quad   &\sum_{i=1}^n T_{ij} \leq q_{j}, \  \forall j \in \{1,\dots,m\}; \ \ \sum_{j=1}^m T_{ij} \leq v_{i}, \ \forall i \in \{1,\dots,n\} \\
    &\sum_{i=1}^n\sum_{j=1}^m T_{ij} = M; \ \ T_{ij} \geq 0, \ \ \forall i \in \{1,\dots,n\}, \forall j \in \{1,\dots,m\}. \\
\end{split}
\label{eq:optTransport}
\end{align}

Optimal transport has been extensively used in data science (\cite{peyre2019computational}), and is a common tool for transfer learning and domain adaptation \cite{courty2014domain}, \cite{courty2017joint}, \cite{lu2017optimal}, \cite{chapel2020partial}. It provides a principled framework to compute a distance between probability distributions, also known as the Wasserstein distance, by finding the minimal effort required to ``transport'' the mass of one distribution to another. Our formulation deviates from the original optimal transport problem by constraining the total mass $M$ to be transported. As we show in Section \ref{sec:partial_alig}, $M$ can be selected in a data-driven fashion, permitting alignments that respect domain-specific regions that are not present in the other domain.   

The user-defined parameters $q_{j}$ and $v_{i}$ indicate the mass assigned to each observation. For instance, to find a hard assignment from each observation in $\Phi_{1}$ to $\Phi_{2}$, and if $n \leq m$, we can set $v_{i} = 1/n$, $q_{j} = 1/n$ and $M = 1$. Soft assignments can be obtained by different choices of masses, or alternatively an entropy regularization $\epsilon \sum
_{i,j} T_{ij}\text{log}(T_{ij})$ can be added to the objective function. In this work we focus on hard assignments since we want to learn one-to-one correspondences. Nevertheless, we state the general formulation, which is useful when there is less confidence in the existence of one-to-one correspondences.

The coupling $T$ contains the information required to combine both manifolds. After a \textit{min-max} normalization denoted by $\tilde{T}$, we can  find a projection of a given sample $x_{i} \in \Phi_{1}$ on $\Phi_{2}$ by its the barycentric projection $x_{i} \mapsto \sum_{j} \tilde{T}_{ij}y_{j}$. Alternatively, we can build a cross-modality similarity matrix $W_{\Phi_{1}\Phi_{2}} = (W_{\Phi_{1}}\tilde{T} + \tilde{T}W_{\Phi_{2}})$, where $W_{\Phi_{k}}$ are the similarities in each domain (computed using Eq. (\ref{eq:DecayGaussKern}) in this paper). Using a similar construction as in \cite{ham2005semisupervised} we can build a joint manifold learning loss: 
\begin{align}
\label{eq:ManifLoss}
\mathcal{L} =   \mu \sum_{ij} ||f_{i} - f_{j}|| W^{ij}_{\Phi_{1}} + \mu \sum_{ij} ||g_{i} - g_{j}|| W^{ij}_{\Phi_{2}} + (1-\mu) \sum_{ij} ||f_{i} - g_{j}||W^{ij}_{\Phi_{1}\Phi_{2}}.
\end{align}
The parameter $\mu$ controls the preservation of the intra-domain geometry. The solution of (\ref{eq:ManifLoss}) provides a shared embedding of the points in both domains and is given by the eigenvectors of the graph Laplacian matrix associated with the joint similarity matrix: 
\begin{align}
\label{eq:w_matrix}
W = 
\left[\begin{array}{cc}
 \mu W_{\Phi_{1}} & (1-\mu) W_{\Phi_{1}\Phi_{2}}\\  
 (1-\mu) W^{'}_{\Phi_{1}\Phi_{2}} &   \mu W_{\Phi_{2}} 
\end{array}
\right].
\end{align}

DTA differs from \cite{ham2005semisupervised} in several ways. First, their method starts by solving (\ref{eq:ManifLoss}), with a $T$ matrix instead of $W_{\Phi_{1}\Phi_{2}}$, which encodes only the \textit{a priori} known correspondences, containing a 1 in entry $(i,j)$ if $x_{i} \in \Phi_{1}$ corresponds to $y_{j} \in \Phi_{2}$ and 0 otherwise. Inter-domain correspondences for the rest of the data are obtained in the latent space produced by the solution. In contrast, DTA first finds a  matrix $T$ that couples all the data, and then builds the inter-domain similarities based on these correspondences. Second, using only $T$ in~(\ref{eq:ManifLoss}) assigns a 0 similarity between $x_{i}$ and the neighbors of $y_{j}$.  We argue that a more natural way to construct the off-diagonal matrices of $W$ is to include the neighbors of $y_{j}$ as being similar to $x_{i}$ as well, motivating our particular construction of $W_{\Phi_{1}\Phi_{2}}$.

\section{Experimental results} 

To demonstrate DTA's effectiveness in finding  a coupling between domains, we compare DTA with  semi-supervised manifold alignment (SSMA) \cite{ham2005semisupervised}, manifold alignment with Procrustes analysis (MA-PA) \cite{wang2008manifold}, and MAGAN \cite{amodio2018magan}. For consistency, we use the same $\alpha$-decay Kernel in Eq.     (\ref{eq:DecayGaussKern}) for the graph-based methods DTA, SSMA, and MA-PA, with $\alpha = 10$ and $k=10$. For MAGAN we use the same architecture provided by the author's code\footnote{https://github.com/KrishnaswamyLab/MAGAN/tree/master/MAGAN}.
MAGAN's architecture is composed of two generators, one mapping from $\Phi_{1}$ to $\Phi_{2}$ and the other in the opposite direction. MAGAN also includes a discriminator for each domain. The model is trained via a min-max game between the generators and discriminators, with a cycle consistency loss \cite{zhu2017unpaired}, and a correspondence loss that tries to preserve the known correspondences. We found that MAGAN usually needs an extra penalization parameter $\rho$ in the correspondence loss to improve its performance, which was not included in the original paper. 

Given the nature of the problem, it is difficult to tune the hyper-parameters present in each method. Thus, we set the same values for each method across all the experiments.  This leave us with one hyperparameter $t$ for DTA, which we set equal to 10 for all the experiments. SSMA and MA-PA require a predefined number of dimensions for the latent space. We set $\rho = 1000$ for MAGAN. A sensitivity analysis for different values is presented in the supplementary material.  

We used four simulated datasets shown in Figure \ref{fig:half_mnist}. \textbf{MNIST-Double}: One domain contains the original MNIST digits, while the other is constructed by downscaling the images to 14x14 pixels, applying a rotation, and adding Gaussian blurring. \textbf{SWISSR-SCURVE}: starting from a common 2D latent space we apply two different transformations resulting in the well known swiss roll and s-curve manifolds embedded in a 3D space. \textbf{Double Helix}: each domain consists of a one dimensional helix plus noise embedded in a 3D space, taken from \cite{tuia2016kernel}. \textbf{SPLATTER-BE}: we simulated single-cell RNA-sequencing data using Splatter \cite{zappia2017splatter}. The difference between $\Phi_{1}$ and $\Phi_{2}$ is due to batch effects, which often arise in biological experiments. For real data, we used the single-cell dataset from the \textit{Multimodal Single-Cell Data Integration} challenge, NeurIPS competition track 2021. The data contains two sets with jointly measured observations for both domains, providing us ground truth information about the coupling between domains. The first set measures gene expression (RNA) and protein abundance (ADT), while the second measures RNA and chromatin accessibility (ATAC). The samples are taken from different donors and batches. We selected  batches ``s1d1'' in both sets for our experiments. Both RNA and ATAC domains are preprocessed, reducing their dimensionality to 1000 features via truncated SVD. 


\subsection{Inter-domain feature mapping}
\label{sub:regression}
Our first comparison metric is the regression performance when mapping between the two domains. When the prior known correspondences are insufficient to successfully train a model, we can improve the training data by expanding the correspondences using each of the considered manifold alignment methods. For DTA, we use hard assignments where for each observation in $\Phi_{1}$ we assign an unique counterpart in $\Phi_{2}$. The correspondences in SSMA and MA-PA are computed as suggested in \cite{ham2005semisupervised} where   the assigned counterpart for each observation in $\Phi_{1}$ corresponds to its nearest sample from $\Phi_{2}$ in the shared latent space. For MAGAN, once the model is trained, we map the data from the first domain into the second using one of the generators. The assigned correspondence is the closest sample. The newly found correspondences serve as the training data for the regression task.

To reduce the dependency on a given regression model, we trained both a fully-connected neural network and a Kernel Ridge Regression (KRR) model. Since the true one-to-one correspondences are accessible to us, the regression models are also trained  with the complete data, as well as the \textit{a priori} known correspondences.  This provides a baseline to show the improvement due to the new information acquired after each of the manifold alignment models, and how well they perform compared to the full correspondence case.

\begin{figure}[!h]
\begin{center}
\centerline{\includegraphics[width = \textwidth]{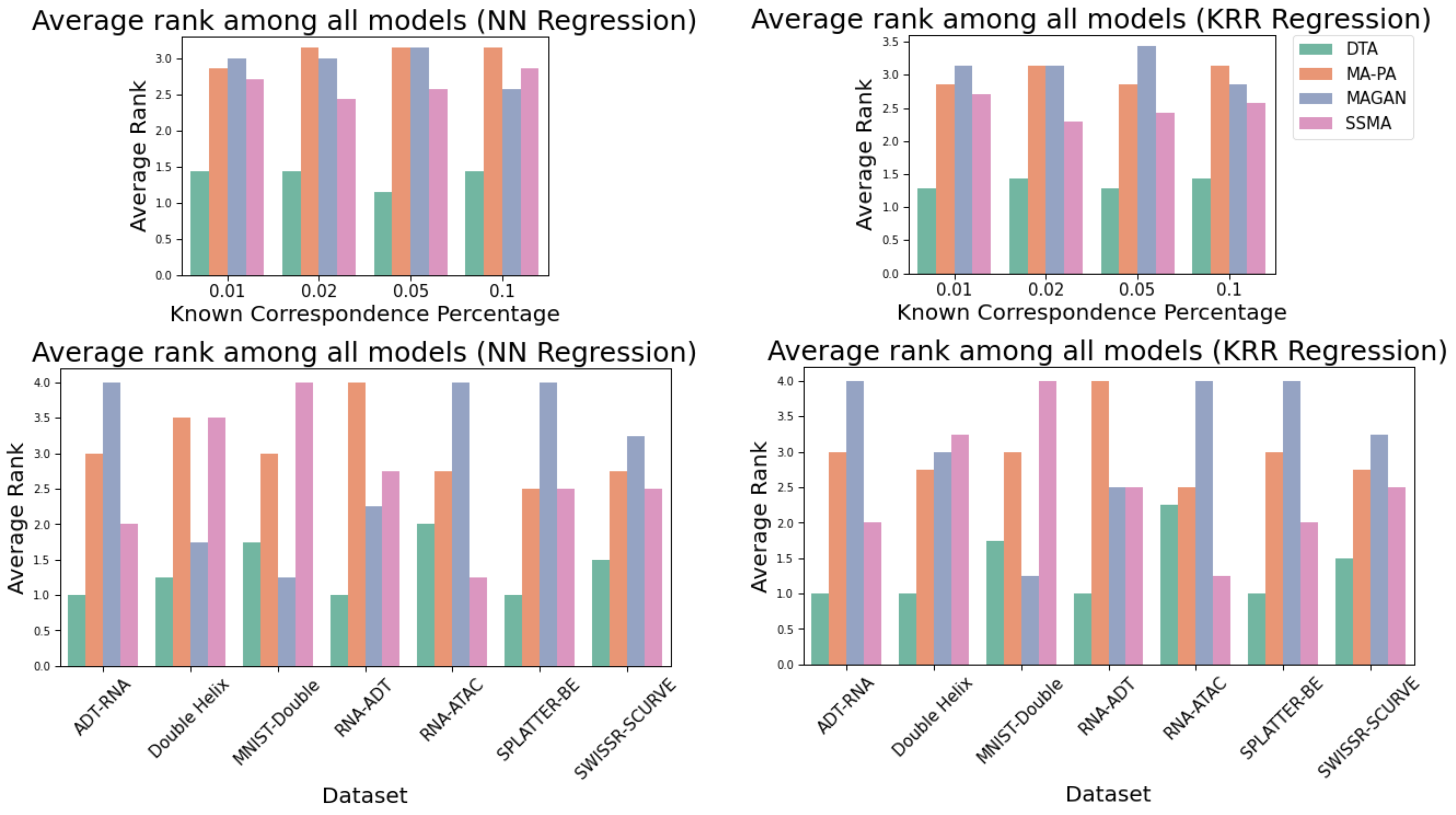}}
\caption{\textbf{Regression results summary.} See the supplement for full results. We computed the rank (1 to 4) of each model in terms of its performance compared with the other three. Top: the average rank of each model, across all datasets for a given known correspondence percentage. Bottom: the average rank across the four known correspondence percentages for a given dataset. DTA almost always outperforms the other methods across all settings.}
\label{fig:reg_results}
\end{center}
\vskip -0.2in
\end{figure}

The results are summarized in Figure \ref{fig:reg_results} with the full test MSE values for each model as well as for the regression trained using all of the correct correspondences are provided in the supplementary material. DTA is the most consistent method as it almost always outperforms the other methods across different datasets and different levels of prior known correspondences. 

\subsection{Domain adaptation}
\label{subsec:domain_adap}
Now we compare the methods on a  domain adaptation problem. Table \ref{tab:classification_results} contains the test error for two $k$-nearest neighbor classifiers, with $k=1$ and $k=10$. The classification models are trained on $\Phi_{2}$ and then tested on the barycentric projections of $\Phi_{1}$. The matrix $\tilde{T}$ is computed for SSMA, MA-PA, and MAGAN from the correspondences as described in Section~\ref{sub:regression}. An alternative approach for SSMA and MA-PA is to train and test the classification on the shared latent representation. For MAGAN the testing can be computed in the generator mapping from $\Phi_{1}$ to $\Phi_{2}$.

\setlength{\tabcolsep}{5pt}
\begin{table}[!h]
\caption{Domain adaptation classification accuracy results under different correspondence percentages. Overall DTA achieves the best results as it is consistently in the top two.}
\label{tab:classification_results}
\centering
\scriptsize
\begin{tabular}{|llllllllll|}
\hline
            & {} & \multicolumn{4}{c}{\underline{KNN-1}} & \multicolumn{4}{c}{\underline{KNN-10}} \vline \\
            &   &                              1\% &                              2\% &                              5\% &                             10\% &                           1\% &                              2\% &                              5\% &                             10\% \\
DATASET & MODEL &                                   &                                   &                                   &                                   &                                   &                                   &                                   &                                   \\
\hline
\multirow{4}{*}{ADT-RNA} & DTA &  {\underline{\textbf{0.672}}} (1) &  {\underline{\textbf{0.712}}} (1) &  {\underline{\textbf{0.719}}} (1) &  {\underline{\textbf{0.725}}} (1) &                \textbf{0.613} (2) &                \textbf{0.647} (2) &                \textbf{0.655} (2) &                \textbf{0.660} (2) \\
            & MA-PA &                         0.539 (3) &                         0.606 (3) &                         0.622 (3) &                         0.642 (3) &                         0.542 (3) &                         0.612 (3) &                         0.626 (3) &                         0.658 (3) \\
            & MAGAN &                         0.473 (4) &                         0.542 (4) &                         0.598 (4) &                         0.590 (4) &                         0.459 (4) &                         0.527 (4) &                         0.586 (4) &                         0.579 (4) \\
            & SSMA &                \textbf{0.633} (2) &                \textbf{0.670} (2) &                \textbf{0.711} (2) &                \textbf{0.722} (2) &  {\underline{\textbf{0.643}}} (1) &  {\underline{\textbf{0.675}}} (1) &  {\underline{\textbf{0.720}}} (1) &  {\underline{\textbf{0.745}}} (1) \\
\hline
\multirow{4}{*}{MNIST-Double} & DTA &                \textbf{0.838} (2) &                \textbf{0.885} (2) &                \textbf{0.905} (2) &                \textbf{0.910} (2) &                \textbf{0.819} (2) &                \textbf{0.852} (2) &                \textbf{0.873} (2) &                \textbf{0.877} (2) \\
            & MA-PA &                         0.679 (3) &                         0.757 (3) &                         0.822 (3) &                         0.841 (3) &                         0.668 (3) &                         0.733 (3) &                         0.791 (3) &                         0.806 (3) \\
            & MAGAN &  {\underline{\textbf{0.960}}} (1) &  {\underline{\textbf{0.939}}} (1) &  {\underline{\textbf{0.952}}} (1) &  {\underline{\textbf{0.987}}} (1) &  {\underline{\textbf{0.895}}} (1) &  {\underline{\textbf{0.884}}} (1) &  {\underline{\textbf{0.894}}} (1) &  {\underline{\textbf{0.906}}} (1) \\
            & SSMA &                         0.493 (4) &                         0.577 (4) &                         0.703 (4) &                         0.820 (4) &                         0.496 (4) &                         0.577 (4) &                         0.701 (4) &                         0.802 (4) \\
\hline
\multirow{4}{*}{RNA-ADT} & DTA &  {\underline{\textbf{0.681}}} (1) &  {\underline{\textbf{0.700}}} (1) &  {\underline{\textbf{0.727}}} (1) &  {\underline{\textbf{0.726}}} (1) &  {\underline{\textbf{0.674}}} (1) &  {\underline{\textbf{0.692}}} (1) &  {\underline{\textbf{0.711}}} (1) &  {\underline{\textbf{0.708}}} (1) \\
            & MA-PA &                         0.563 (4) &                         0.625 (4) &                         0.660 (4) &                         0.666 (4) &                         0.518 (4) &                         0.555 (4) &                         0.591 (4) &                         0.581 (4) \\
            & MAGAN &                         0.605 (3) &                         0.653 (3) &                         0.696 (3) &                         0.673 (3) &                \textbf{0.602} (2) &                \textbf{0.651} (2) &                \textbf{0.668} (2) &                \textbf{0.675} (2) \\
            & SSMA &                \textbf{0.649} (2) &                \textbf{0.670} (2) &                \textbf{0.703} (2) &                \textbf{0.721} (2) &                         0.599 (3) &                         0.615 (3) &                         0.655 (3) &                         0.659 (3) \\
\hline
\multirow{4}{*}{RNA-ATAC} & DTA &  {\underline{\textbf{0.658}}} (1) &                \textbf{0.696} (2) &                         0.718 (3) &                \textbf{0.722} (2) &  {\underline{\textbf{0.617}}} (1) &  {\underline{\textbf{0.643}}} (1) &  {\underline{\textbf{0.673}}} (1) &                \textbf{0.678} (2) \\
            & MA-PA &                         0.574 (3) &  {\underline{\textbf{0.705}}} (1) &                \textbf{0.758} (2) &                         0.683 (3) &                         0.534 (3) &                         0.615 (3) &                         0.637 (3) &                         0.596 (3) \\
            & MAGAN &                         0.286 (4) &                         0.301 (4) &                         0.422 (4) &                         0.519 (4) &                         0.296 (4) &                         0.305 (4) &                         0.435 (4) &                         0.527 (4) \\
            & SSMA &                \textbf{0.642} (2) &                         0.687 (3) &  {\underline{\textbf{0.780}}} (1) &  {\underline{\textbf{0.807}}} (1) &                \textbf{0.603} (2) &                \textbf{0.621} (2) &                \textbf{0.670} (2) &  {\underline{\textbf{0.691}}} (1) \\
\hline
\multirow{4}{*}{SPLATTER-BE} & DTA &  {\underline{\textbf{0.684}}} (1) &  {\underline{\textbf{0.743}}} (1) &  {\underline{\textbf{0.773}}} (1) &  {\underline{\textbf{0.761}}} (1) &  {\underline{\textbf{0.660}}} (1) &  {\underline{\textbf{0.714}}} (1) &  {\underline{\textbf{0.742}}} (1) &  {\underline{\textbf{0.735}}} (1) \\
            & MA-PA &                \textbf{0.546} (2) &                         0.526 (3) &                         0.530 (3) &                         0.532 (3) &                \textbf{0.551} (2) &                         0.527 (3) &                         0.532 (3) &                         0.535 (3) \\
            & MAGAN &                         0.266 (4) &                         0.341 (4) &                         0.417 (4) &                         0.486 (4) &                         0.260 (4) &                         0.348 (4) &                         0.420 (4) &                         0.491 (4) \\
            & SSMA &                         0.485 (3) &                \textbf{0.589} (2) &                \textbf{0.677} (2) &                \textbf{0.709} (2) &                         0.485 (3) &                \textbf{0.590} (2) &                \textbf{0.674} (2) &                \textbf{0.708} (2) \\
\hline
\end{tabular}

\end{table}

Overall, DTA achieves the best results. MAGAN performs considerably better for MNIST-Double as in Section~\ref{sub:regression}, but it tends to have the worst performance in the more complex single-cell datasets. In general the graph-based methods have a similar performance for ADT-RNA and the opposite direction (RNA-ADT), but MAGAN has a drastically different performance when ADT is mapped to RNA than when RNA is mapped to ADT. 

\subsection{Partial alignment}
\label{sec:partial_alig}

\begin{figure}[!h]
\begin{center}
\centerline{\includegraphics[width = \textwidth]{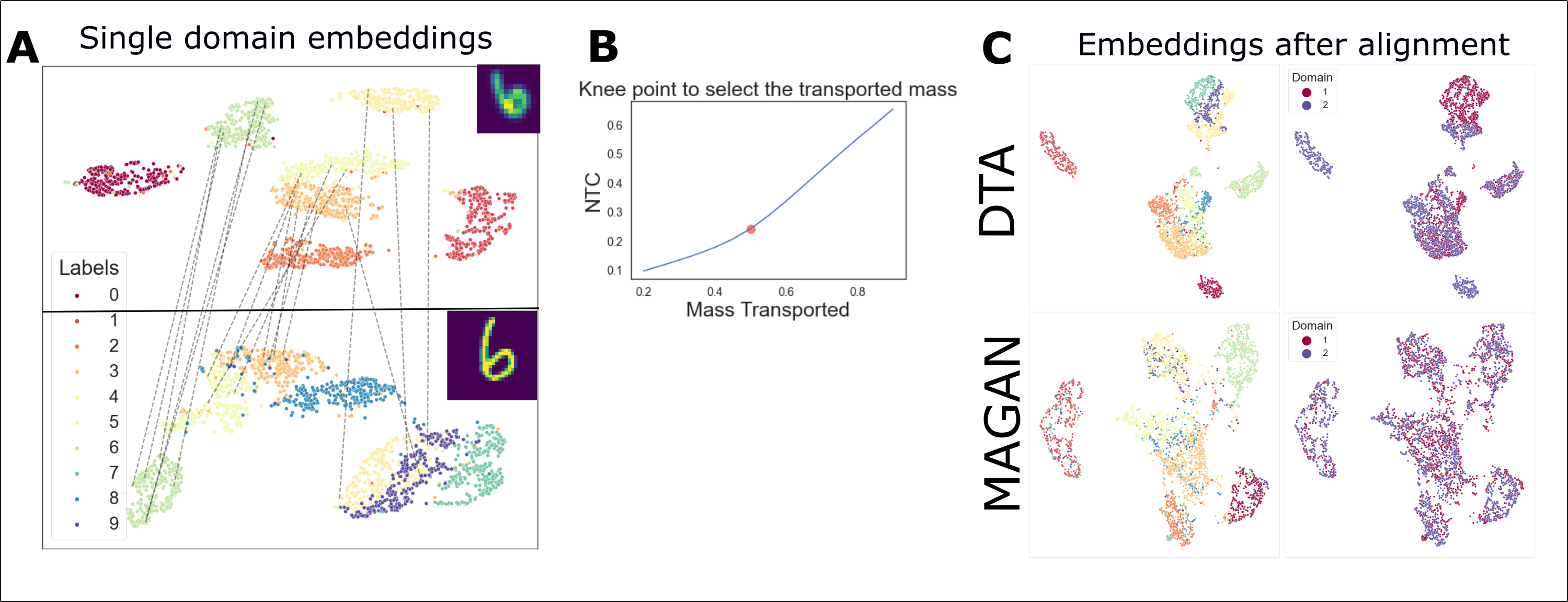}}
\caption{\textbf{Partial alignment.} We subset both domains of the MNIST-Double dataset such that both domains contain specific regions with no counterpart in the other domain. \textbf{A)} Domain specific 2D UMAP (\cite{mcinnes2018umap}) embeddings and dashed lines connecting the \textit{a priori} known correspondences. \textbf{B)} Knee plot used to indentify the optimal mass $M$ to be transported. \textbf{C)} Joint embedding of both domains after alignment, colored by labels and domain membership. DTA is able to retain domain-specific regions separate, while combining successfully the true counterparts. In contrast, MAGAN maps regions of $\Phi_{1}$ to non-corresponding counterparts in $\Phi_{2}$.}
\label{fig:part_align}
\end{center}
\vskip -0.2in
\end{figure}

Here we show the ability of DTA to perform partial alignment. Figure \ref{fig:part_align} conceptualizes this scenario where the data in one or both domains is not completely represented in the other. If, for instance, we use MAGAN to perform the alignment, the nature of its min-max game will map samples from one domain into high density regions of the other. This causes false positive correspondences, and an incorrect alignment for some portions of the data. In contrast, DTA can handle this scenario in a data-driven way. The idea is to select a value of $M$ in problem (\ref{eq:optTransport}), that corresponds to the mass from $\Phi_{1}$ that has an actual counterpart in $\Phi_{2}$. We select $M$ using the normalized transportation cost:
\begin{align}
    NTC = \frac{\sum_{ij}D_{ij}T_{ij}}{M}
\end{align}

After selecting a grid of values for $M$ ranging from 0 to 1, we solve (\ref{eq:optTransport}) for each particular value and compute its corresponding $NTC$. The transportation cost for observations far away from the known correspondences (i.e. points that are present in only one of the domains) starts to increase  rapidly after a certain threshold that likely corresponds to the case where all of the shared points have been aligned. Thus the selected mass $M$ to be transported is computed by identifying a knee point in the $NTC$ vs $M$ plot (Figure~\ref{fig:part_align}B).

A quantitative evaluation  of DTA and MAGAN in this scenario is presented in Table \ref{tab:Pclassification_results}. After finding the \textit{min-max} normalized coupling matrix $T$, we compute $W$ via (\ref{eq:w_matrix}) and transform it to a distance matrix used in a knn classifier. The test accuracy values are reported and, as expected, the results show how MAGAN maps observations close to incompatible regions on $\Phi_{2}$, deteriorating the performance of the classifier. 

\setlength{\tabcolsep}{4.5pt}
\begin{table}[H]
\caption{Test accuracy for the partial alignment experiments. DTA outperforms MAGAN.}
\label{tab:Pclassification_results}
\centering
\scriptsize
\begin{tabular}{|llllllllll|}
\hline
            & {} & \multicolumn{4}{c}{\underline{KNN-1}} & \multicolumn{4}{c}{\underline{KNN-10}} \vline \\
            &   &                              1\% &                              2\% &                              5\% &                             10\% &                           1\% &                              2\% &                              5\% &                             10\% \\
DATASET & MODEL &                                   &                                   &                                   &                                   &                                   &                                   &                                   &                                   \\
\hline
\multirow{2}{*}{MNIST-Double (P)} & DTA &  {\underline{\textbf{0.821}}} &  {\underline{\textbf{0.861}}} &  {\underline{\textbf{0.882}}}  &                {\underline{\textbf{0.887}}}  &  {\underline{\textbf{0.900}}}  &  {\underline{\textbf{0.917}}}  &  {\underline{\textbf{0.924}}}  &  {\underline{\textbf{0.926}}}  \\
            & MAGAN &                         0.583  &                         0.663  &                         0.720  &                         0.743  &                         0.753  &                         0.801  &                         0.827  &                         0.836  \\
\hline
\multirow{2}{*}{RNA-ADT (P)} & DTA &                {\underline{\textbf{0.820}}}  &                {\underline{\textbf{0.831}}}&  {\underline{\textbf{0.844}}}  &  {\underline{\textbf{0.849}}}  &  {\underline{\textbf{0.910}}}  &  {\underline{\textbf{0.910}}}  &  {\underline{\textbf{0.912}}}  &  {\underline{\textbf{0.919}}}  \\
            & MAGAN &                         0.627  &                         0.655  &                         0.675  &                         0.679  &                         0.692 &                         0.719  &                         0.726  &                         0.726  \\
\hline
\end{tabular}

\end{table}

\subsection{Feature concatenation via DTA}
\label{sub:Feature}

\begin{figure}[!h]
\begin{center}
\centerline{\includegraphics[width =  0.85\textwidth]{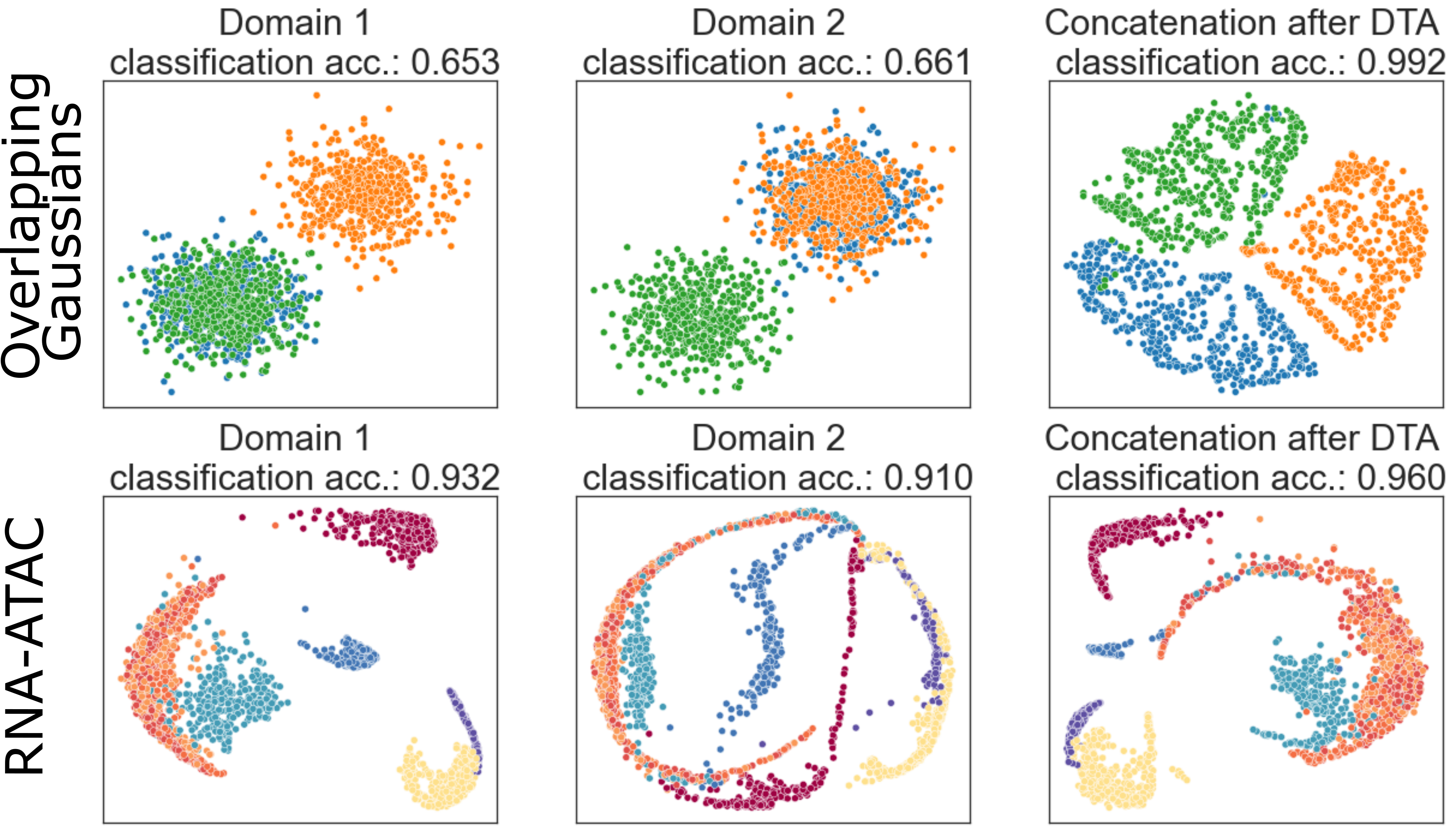}}
\caption{\textbf{Feature concatenation.} The top row shows a toy example where the data consists of three 2D Gaussian distributions. Each domain considered separately is not sufficient to separate the three classes. After applying DTA with only 1\% of known correspondences and Eq.  (\ref{eq:labelInfo}) as the cost matrix, the features of matching observations are concatenated and the new 4D representation is enough to classify the three groups. The 2D projection of the concatenated data is computed with PHATE \cite{moon2019visualizing} and displayed in the third column. The bottom row exhibits a similar procedure but for the real single-cell data (RNA-ATAC). After feature concatenation the classification accuracy is improved in comparison to the cases where the domains are considered separately.}
\label{fig:featureConcat}
\end{center}
\vskip -0.2in
\end{figure}

In some applications, different views may contain information not present in the other. For instance some classes in domain 1 may be heavily overlapped, while they remain separate in the feature space of domain 2 (Figure~\ref{fig:featureConcat}, top). DTA finds an alignment that improves the classification after concatenating the features of assigned correspondences. In this problem, we have access to label information in both domains; e.g., cell types may be identified in both single cell RNA-sequencing and ATAC-sequencing measurements. The label information provides us extra information that  can also be exploited to obtain a better alignment. To do this, we use a 0-1 hot vector encoding scheme for the labels, allowing us to include a ``label based distance".  This results in a modification  of the distances matrix $D$: 
\begin{align}
D^{\mathit{l}}_{ij} = D_{ij} + \mathbbm{1}[i_{l} \neq j_{l}],
\label{eq:labelInfo}
\end{align}
where $\mathbbm{1}[i_{l} \neq j_{l}]$ is an indicator function that checks if the labels for the $i$th point in $\Phi_1$ and the $j$th point in $\Phi_2$  are different.  The coupling matrix $T$ is obtained by solving (\ref{eq:optTransport}) with the new cost matrix $D^{\mathit{l}}$. Figure \ref{fig:featureConcat} demonstrates this application, where we report the  accuracy  using a Random Forest before and after concatenation. Concatenating via DTA improves the accuracy.

\subsection{DTA recovers Mutual Information}
\label{sub:MI}
\begin{figure}[!h]
\begin{center}
\centerline{\includegraphics[width =  0.95\textwidth]{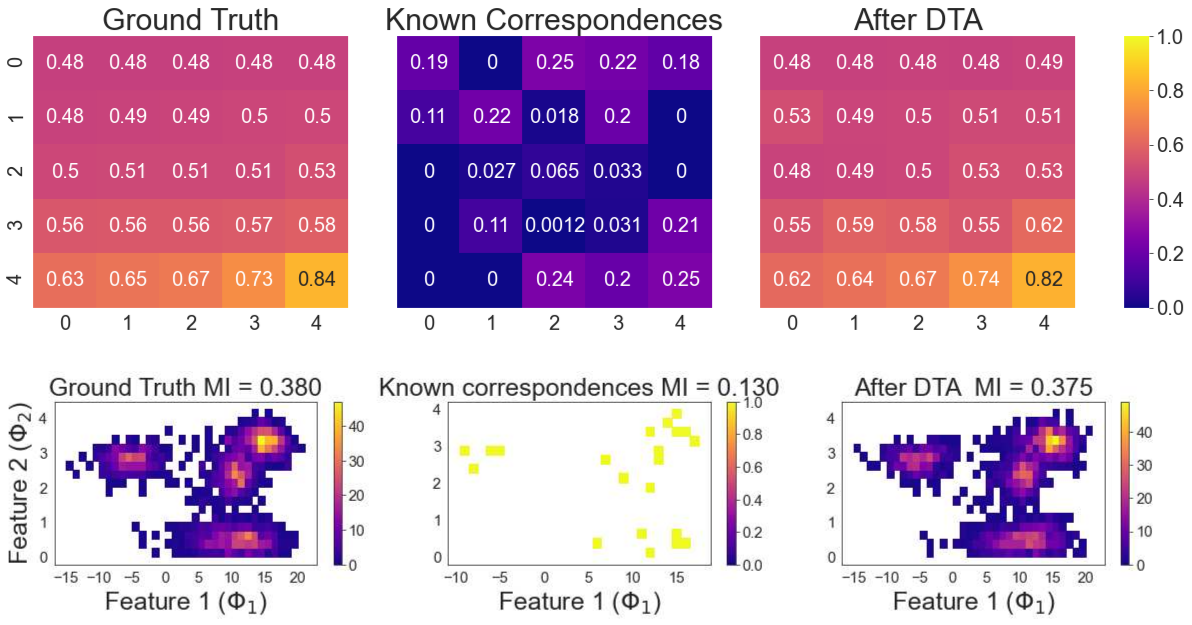}}
\caption{\textbf{Mutual information.} \textbf{A)} The first column shows the highest MI values between 25 pairs of features of the RNA-ADT dataset using the whole data. The estimated values for the same pairs using only the known correspondences are not able to capture the relationship between features. The third column shows the values after applying DTA and recovering a coupling between domains. \textbf{B)} We zoom in and inspect the joint distribution and the MI for a particular pair of features, using the whole data, the known correspondences, and  after DTA.}
\label{fig:MutualInf}
\end{center}
\vskip -0.2in
\end{figure}

Finally we show how the inter-domain correspondences learned by DTA accurately recover the mutual information (MI) between pairs of features from the two domains (Figure \ref{fig:MutualInf}). For this experiment we compute the MI between the features of RNA and ADT using the ground truth pairing, and we keep the 25 pairs with the highest value.  Assuming we only have access to a small subset of matching pairs between domains (known correspondences) the MI estimate is inaccurate. But after applying DTA, the estimated MI is much closer to the true value.

\section{Conclusion and Limitations}

We introduced Diffusion Transport Alignment (DTA), a manifold alignment method that exploits prior known correspondences between two related domains. We showed that DTA is superior to previous state-of-the-art  manifold alignment methods by various metrics of comparison. DTA is able to recover meaningful connections that can be leveraged for downstream analysis tasks that may be otherwise difficult to perform. We also showed that partial manifold alignment can be handled by DTA, reducing the likelihood of falsely connecting points between domains, whereas previous methods are not naturally equipped to tackle this case.

DTA is a graph-based method that relies on the particular notion of similarity imposed on the data. It tends to match observations between domains that share a similar connection to the known correspondences. Thus, differences in domain density could affect its performance. This issue also affects SSMA and MA-PA, whereas MAGAN seems to be more resilient to density differences. The nature of the inter-domain connections via diffusion makes the optimal transport problem prefer to assign correspondences between observations close to the known correspondences. If a given observation is the closest neighbor to a correspondence sample in one domain, but due to density differences it does not belong to its neighbors in the other domain,  DTA will produce a wrong correspondence. Also, DTA as well as the other graph-based methods (SSMA and MA-PA), do not scale as efficiently to larger datasets as a neural network approach such as MAGAN. Mitigation strategies to reduce these limitations are left as future work.


\newpage
\bibliography{ref}

\begin{thebibliography}{10}

\bibitem{amodio2018magan}
M.~Amodio and S.~Krishnaswamy.
\newblock Magan: Aligning biological manifolds.
\newblock In {\em International Conference on Machine Learning}, pages
  215--223. PMLR, 2018.

\bibitem{belkin2003laplacian}
M.~Belkin and P.~Niyogi.
\newblock Laplacian eigenmaps for dimensionality reduction and data
  representation.
\newblock {\em Neural computation}, 15(6):1373--1396, 2003.

\bibitem{cao2020unsupervised}
K.~Cao, X.~Bai, Y.~Hong, and L.~Wan.
\newblock Unsupervised topological alignment for single-cell multi-omics
  integration.
\newblock {\em Bioinformatics}, 36(Supplement\_1):i48--i56, 2020.

\bibitem{cao2022manifold}
K.~Cao, Y.~Hong, and L.~Wan.
\newblock Manifold alignment for heterogeneous single-cell multi-omics data
  integration using pamona.
\newblock {\em Bioinformatics}, 38(1):211--219, 2022.

\bibitem{chang2013canonical}
B.~Chang, U.~Kruger, R.~Kustra, and J.~Zhang.
\newblock Canonical correlation analysis based on hilbert-schmidt independence
  criterion and centered kernel target alignment.
\newblock In {\em International Conference on Machine Learning}, pages
  316--324. PMLR, 2013.

\bibitem{chapel2020partial}
L.~Chapel, M.~Z. Alaya, and G.~Gasso.
\newblock Partial optimal transport with applications on positive-unlabeled
  learning.
\newblock {\em arXiv preprint arXiv:2002.08276}, 2020.

\bibitem{coifman2006diffusion}
R.~R. Coifman and S.~Lafon.
\newblock Diffusion maps.
\newblock {\em Applied and computational harmonic analysis}, 21(1):5--30, 2006.

\bibitem{courty2017joint}
N.~Courty, R.~Flamary, A.~Habrard, and A.~Rakotomamonjy.
\newblock Joint distribution optimal transportation for domain adaptation.
\newblock {\em Advances in Neural Information Processing Systems}, 30, 2017.

\bibitem{courty2014domain}
N.~Courty, R.~Flamary, and D.~Tuia.
\newblock Domain adaptation with regularized optimal transport.
\newblock In {\em Joint European Conference on Machine Learning and Knowledge
  Discovery in Databases}, pages 274--289. Springer, 2014.

\bibitem{cui2014generalized}
Z.~Cui, H.~Chang, S.~Shan, and X.~Chen.
\newblock Generalized unsupervised manifold alignment.
\newblock {\em Advances in Neural Information Processing Systems}, 27, 2014.

\bibitem{demetci2022scot}
P.~Demetci, R.~Santorella, B.~Sandstede, W.~S. Noble, and R.~Singh.
\newblock Scot: Single-cell multi-omics alignment with optimal transport.
\newblock {\em Journal of Computational Biology}, 29(1):3--18, 2022.

\bibitem{gao2012multi}
G.~Gao and H.~Ma.
\newblock Multi-modality movie scene detection using kernel canonical
  correlation analysis.
\newblock In {\em Proceedings of the 21st International Conference on Pattern
  Recognition (ICPR2012)}, pages 3074--3077. IEEE, 2012.

\bibitem{gravina2017multi}
R.~Gravina, P.~Alinia, H.~Ghasemzadeh, and G.~Fortino.
\newblock Multi-sensor fusion in body sensor networks: State-of-the-art and
  research challenges.
\newblock {\em Information Fusion}, 35:68--80, 2017.

\bibitem{ham2005semisupervised}
J.~Ham, D.~Lee, and L.~Saul.
\newblock Semisupervised alignment of manifolds.
\newblock In {\em International Workshop on Artificial Intelligence and
  Statistics}, pages 120--127. PMLR, 2005.

\bibitem{ham2003learning}
J.~H. Ham, D.~D. Lee, and L.~K. Saul.
\newblock Learning high dimensional correspondences from low dimensional
  manifolds.
\newblock 2003.

\bibitem{hu2019mima}
J.~Hu, D.~Hong, and X.~X. Zhu.
\newblock Mima: Mapper-induced manifold alignment for semi-supervised fusion of
  optical image and polarimetric sar data.
\newblock {\em IEEE Transactions on Geoscience and Remote Sensing},
  57(11):9025--9040, 2019.

\bibitem{katz2019alternating}
O.~Katz, R.~Talmon, Y.-L. Lo, and H.-T. Wu.
\newblock Alternating diffusion maps for multimodal data fusion.
\newblock {\em Information Fusion}, 45:346--360, 2019.

\bibitem{kuchroo2021multimodal}
M.~Kuchroo, A.~Godavarthi, A.~Tong, G.~Wolf, and S.~Krishnaswamy.
\newblock Multimodal data visualization and denoising with integrated
  diffusion.
\newblock In {\em 2021 IEEE 31st International Workshop on Machine Learning for
  Signal Processing (MLSP)}, pages 1--6. IEEE, 2021.

\bibitem{lafon2006data}
S.~Lafon, Y.~Keller, and R.~R. Coifman.
\newblock Data fusion and multicue data matching by diffusion maps.
\newblock {\em IEEE Transactions on pattern analysis and machine intelligence},
  28(11):1784--1797, 2006.

\bibitem{lahat2015multimodal}
D.~Lahat, T.~Adali, and C.~Jutten.
\newblock Multimodal data fusion: an overview of methods, challenges, and
  prospects.
\newblock {\em Proceedings of the IEEE}, 103(9):1449--1477, 2015.

\bibitem{lindenbaum2020multi}
O.~Lindenbaum, A.~Yeredor, M.~Salhov, and A.~Averbuch.
\newblock Multi-view diffusion maps.
\newblock {\em Information Fusion}, 55:127--149, 2020.

\bibitem{liu2016manifold}
Z.~Liu, W.~Wang, and Q.~Jin.
\newblock Manifold alignment using discrete surface ricci flow.
\newblock {\em CAAI Transactions on Intelligence Technology}, 1(3):285--292,
  2016.

\bibitem{lu2017optimal}
Y.~Lu, L.~Chen, and A.~Saidi.
\newblock Optimal transport for deep joint transfer learning.
\newblock {\em arXiv preprint arXiv:1709.02995}, 2017.

\bibitem{mcinnes2018umap}
L.~McInnes, J.~Healy, and J.~Melville.
\newblock Umap: Uniform manifold approximation and projection for dimension
  reduction.
\newblock {\em arXiv preprint arXiv:1802.03426}, 2018.

\bibitem{moon2019visualizing}
K.~R. Moon, D.~van Dijk, Z.~Wang, S.~Gigante, D.~B. Burkhardt, W.~S. Chen,
  K.~Yim, A.~van~den Elzen, M.~J. Hirn, R.~R. Coifman, et~al.
\newblock Visualizing structure and transitions in high-dimensional biological
  data.
\newblock {\em Nature biotechnology}, 37(12):1482--1492, 2019.

\bibitem{peyre2019computational}
G.~Peyr{\'e}, M.~Cuturi, et~al.
\newblock Computational optimal transport: With applications to data science.
\newblock {\em Foundations and Trends{\textregistered} in Machine Learning},
  11(5-6):355--607, 2019.

\bibitem{stanley2020harmonic}
J.~S. Stanley~III, S.~Gigante, G.~Wolf, and S.~Krishnaswamy.
\newblock Harmonic alignment.
\newblock In {\em Proceedings of the 2020 SIAM International Conference on Data
  Mining}, pages 316--324. SIAM, 2020.

\bibitem{stuart2019comprehensive}
T.~Stuart, A.~Butler, P.~Hoffman, C.~Hafemeister, E.~Papalexi, W.~M. Mauck~III,
  Y.~Hao, M.~Stoeckius, P.~Smibert, and R.~Satija.
\newblock Comprehensive integration of single-cell data.
\newblock {\em Cell}, 177(7):1888--1902, 2019.

\bibitem{thompson1984canonical}
B.~Thompson.
\newblock {\em Canonical correlation analysis: Uses and interpretation}.
\newblock Number~47. Sage, 1984.

\bibitem{tuia2016kernel}
D.~Tuia and G.~Camps-Valls.
\newblock Kernel manifold alignment for domain adaptation.
\newblock {\em PloS one}, 11(2):e0148655, 2016.

\bibitem{vieira2020multimodal}
S.~Vieira, W.~H.~L. Pinaya, R.~Garcia-Dias, and A.~Mechelli.
\newblock Multimodal integration.
\newblock In {\em Machine Learning}, pages 283--305. Elsevier, 2020.

\bibitem{wang2008manifold}
C.~Wang and S.~Mahadevan.
\newblock Manifold alignment using procrustes analysis.
\newblock In {\em Proceedings of the 25th international conference on Machine
  learning}, pages 1120--1127, 2008.

\bibitem{wang2009manifold}
C.~Wang and S.~Mahadevan.
\newblock Manifold alignment without correspondence.
\newblock In {\em Twenty-First International Joint Conference on Artificial
  Intelligence}, 2009.

\bibitem{wang2011heterogeneous}
C.~Wang and S.~Mahadevan.
\newblock Heterogeneous domain adaptation using manifold alignment.
\newblock In {\em Twenty-second international joint conference on artificial
  intelligence}, 2011.

\bibitem{zappia2017splatter}
L.~Zappia, B.~Phipson, and A.~Oshlack.
\newblock Splatter: simulation of single-cell rna sequencing data.
\newblock {\em Genome biology}, 18(1):1--15, 2017.

\bibitem{zhu2017unpaired}
J.-Y. Zhu, T.~Park, P.~Isola, and A.~A. Efros.
\newblock Unpaired image-to-image translation using cycle-consistent
  adversarial networks.
\newblock In {\em Proceedings of the IEEE international conference on computer
  vision}, pages 2223--2232, 2017.

\end{thebibliography}
\bibliographystyle{abbrv}

\newpage

\appendix

\section{Supplementary material}

\subsection{Regression results}

\begin{figure}[!h]
\begin{center}
\centerline{\includegraphics[width = 1.2\textwidth]{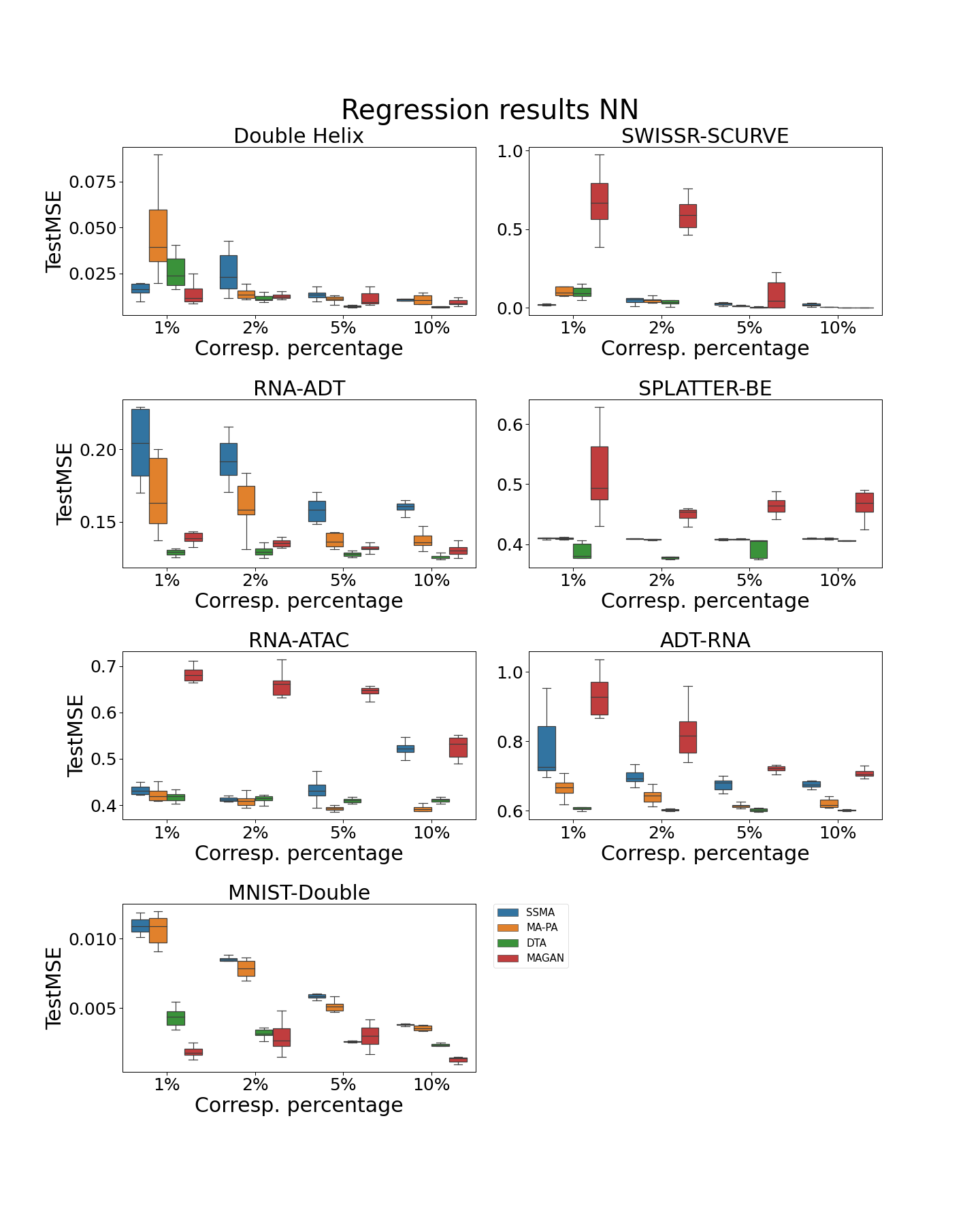}}
\caption{\textbf{Regression boxplots for the Neural Network model.} The boxplots are created from 10 randomized runs for each model and for all of the datasets. See Table~\ref{tab:metrics_manifold} for the average MSE values and Figure~\ref{fig:reg_results} for a summary across datasets and across correspondence percentages.}
\label{fig:reg_boxplots1}
\end{center}
\vskip -0.2in
\end{figure}

Here we provide further details of the results in Section \ref{sub:regression}. Table \ref{tab:metrics_manifold} contains the mean over 10 runs for each model across the different datasets assuming different percentage levels of prior known correspondences. We include the results using all the data (AllData) which is considered the ``best'' possible result, since it contains all ground truth matchings between domains. On the other hand, if the models are trained by only employing the prior known correspondences (PriorInfo), we expect to see the worst results. Thus each of the manifold alignment models should improve upon the PriorInfo results after recovering more correspondences. More details about the variance of the results are displayed in Figure \ref{fig:reg_boxplots1}. 
The results show that DTA consistently performs the best or close to it. No other method is as consistent in its performance as DTA. This is all corroborated in Figure~\ref{fig:reg_results}, which  summarizes the results.

\addtolength{\tabcolsep}{-3pt}
\begin{table}[H]
\caption{Full regression results. Overall, DTA is the most consistent method as summarized in Figure~\ref{fig:reg_results}.}
\label{tab:metrics_manifold}
\centering
\scriptsize
\begin{tabular}{|llcccccccc|}
\hline
              & {} & \multicolumn{4}{c}{Test MSE (Neural Network)} & \multicolumn{4}{c}{Test MSE (KRR)} \vline \\
              & Correspondence percentage &                              1\% &                              2\% &                             5\% &                            10\% &                             1\% &                              2\% &                             5\% &                            10\% \\
Dataset & Model &                                   &                                   &                                   &                                   &                                   &                                   &                                   &                                   \\
\hline
\multirow{6}{*}{ADT-RNA} & AllData &   0.580 &  0.581 &  0.580 &  0.582 &      0.585 &  0.585 &  0.585 &  0.586 \\
                & PriorInfo &   0.791 &  0.711 &  0.666 &  0.647 &      0.746 &  0.683 &  0.638 &  0.614
               \\
              & DTA &  {\underline{\textbf{0.612}}} (1) &  {\underline{\textbf{0.603}}} (1) &  {\underline{\textbf{0.603}}} (1) &  {\underline{\textbf{0.602}}} (1) &  {\underline{\textbf{0.636}}} (1) &  {\underline{\textbf{0.614}}} (1) &  {\underline{\textbf{0.613}}} (1) &  {\underline{\textbf{0.612}}} (1) \\
              & MA-PA &                         0.777 (3) &                         0.698 (3) &                         0.677 (3) &                         0.682 (3) &                         0.896 (3) &                         0.760 (3) &                         0.706 (3) &                         0.692 (3) \\
              & MAGAN &                         0.952 (4) &                         0.837 (4) &                         0.724 (4) &                         0.707 (4) &                         0.980 (4) &                         0.859 (4) &                         0.730 (4) &                         0.700 (4) \\
              & SSMA &                \textbf{0.670} (2) &                \textbf{0.643} (2) &                \textbf{0.615} (2) &                \textbf{0.621} (2) &                \textbf{0.691} (2) &                \textbf{0.662} (2) &                \textbf{0.626} (2) &                \textbf{0.629} (2) \\
\hline
\multirow{6}{*}{Double Helix} & AllData &   0.007 &  0.008 &  0.006 &  0.006 &      0.132 &  0.132 &  0.132 &  0.132 \\
& PriorInfo &   0.107 &  0.023 &  0.010 &  0.008 &      0.163 &  0.145 &  0.138 &  0.137 \\
              & DTA &                \textbf{0.029} (2) &  {\underline{\textbf{0.012}}} (1) &  {\underline{\textbf{0.007}}} (1) &  {\underline{\textbf{0.007}}} (1) &  {\underline{\textbf{0.133}}} (1) &  {\underline{\textbf{0.132}}} (1) &  {\underline{\textbf{0.132}}} (1) &  {\underline{\textbf{0.132}}} (1) \\
              & MA-PA &                         0.031 (3) &                         0.026 (4) &                         0.014 (4) &                         0.010 (3) &                         0.134 (3) &                         0.137 (4) &                \textbf{0.133} (2) &                \textbf{0.133} (2) \\
              & MAGAN &  {\underline{\textbf{0.014}}} (1) &                \textbf{0.013} (2) &                \textbf{0.011} (2) &                \textbf{0.009} (2) &                \textbf{0.134} (2) &                         0.133 (3) &                         0.134 (4) &                         0.134 (3) \\
              & SSMA &                         0.053 (4) &                         0.014 (3) &                         0.011 (3) &                         0.011 (4) &                         0.145 (4) &                \textbf{0.133} (2) &                         0.133 (3) &                         0.134 (4) \\
\hline
\multirow{6}{*}{MNIST-Double} & AllData &   0.000 &  0.000 &  0.000 &  0.000 &      0.000 &  0.000 &  0.000 &  0.000 \\
                & PriorInfo &   0.010 &  0.006 &  0.002 &  0.001 &      0.008 &  0.004 &  0.001 &  0.000 \\
              & DTA &                \textbf{0.004} (2) &                \textbf{0.003} (2) &  {\underline{\textbf{0.003}}} (1) &                \textbf{0.002} (2) &                \textbf{0.004} (2) &                \textbf{0.003} (2) &  {\underline{\textbf{0.002}}} (1) &                \textbf{0.002} (2) \\
              & MA-PA &                         0.011 (3) &                         0.008 (3) &                         0.005 (3) &                         0.004 (3) &                         0.010 (3) &                         0.007 (3) &                         0.005 (3) &                         0.003 (3) \\
              & MAGAN &  {\underline{\textbf{0.002}}} (1) &  {\underline{\textbf{0.003}}} (1) &                \textbf{0.003} (2) &  {\underline{\textbf{0.001}}} (1) &  {\underline{\textbf{0.001}}} (1) &  {\underline{\textbf{0.002}}} (1) &                \textbf{0.003} (2) &  {\underline{\textbf{0.001}}} (1) \\
              & SSMA &                         0.011 (4) &                         0.009 (4) &                         0.006 (4) &                         0.004 (4) &                         0.011 (4) &                         0.008 (4) &                         0.005 (4) &                         0.003 (4) \\
\hline
\multirow{6}{*}{RNA-ADT} & AllData &   0.109 &  0.108 &  0.109 &  0.109 &      0.104 &  0.104 &  0.104 &  0.105 \\ 
& PriorInfo &   0.715 &  0.528 &  0.334 &  0.240 &      0.354 &  0.215 &  0.174 &  0.173\\
              & DTA &  {\underline{\textbf{0.130}}} (1) &  {\underline{\textbf{0.130}}} (1) &  {\underline{\textbf{0.128}}} (1) &  {\underline{\textbf{0.126}}} (1) &  {\underline{\textbf{0.116}}} (1) &  {\underline{\textbf{0.115}}} (1) &  {\underline{\textbf{0.113}}} (1) &  {\underline{\textbf{0.113}}} (1) \\
              & MA-PA &                         0.219 (4) &                         0.192 (4) &                         0.158 (4) &                         0.160 (4) &                         0.223 (4) &                         0.181 (4) &                         0.134 (4) &                         0.134 (4) \\
              & MAGAN &                         0.187 (3) &                \textbf{0.135} (2) &                \textbf{0.132} (2) &                \textbf{0.131} (2) &                         0.174 (3) &                \textbf{0.123} (2) &                \textbf{0.121} (2) &                         0.121 (3) \\
              & SSMA &                \textbf{0.169} (2) &                         0.161 (3) &                         0.141 (3) &                         0.137 (3) &                \textbf{0.147} (2) &                         0.140 (3) &                         0.121 (3) &                \textbf{0.118} (2) \\
\hline
\multirow{6}{*}{RNA-ATAC} & AllData &   0.369 &  0.369 &  0.369 &  0.370 &      0.346 &  0.346 &  0.347 &  0.346 \\
& PriorInfo &   0.497 &  0.472 &  0.435 &  0.398 &      0.397 &  0.381 &  0.362 &  0.354 \\
              & DTA &  {\underline{\textbf{0.418}}} (1) &                         0.414 (3) &                \textbf{0.410} (2) &                \textbf{0.410} (2) &  {\underline{\textbf{0.419}}} (1) &                         0.416 (3) &                         0.408 (3) &                \textbf{0.410} (2) \\
              & MA-PA &                         0.433 (3) &                \textbf{0.413} (2) &                         0.433 (3) &                         0.517 (3) &                         0.462 (3) &                \textbf{0.410} (2) &                \textbf{0.404} (2) &                         0.495 (3) \\
              & MAGAN &                         0.677 (4) &                         0.661 (4) &                         0.645 (4) &                         0.525 (4) &                         0.678 (4) &                         0.658 (4) &                         0.640 (4) &                         0.520 (4) \\
              & SSMA &                \textbf{0.424} (2) &  {\underline{\textbf{0.413}}} (1) &  {\underline{\textbf{0.393}}} (1) &  {\underline{\textbf{0.393}}} (1) &                \textbf{0.422} (2) &  {\underline{\textbf{0.409}}} (1) &  {\underline{\textbf{0.380}}} (1) &  {\underline{\textbf{0.379}}} (1) \\
\hline
\multirow{6}{*}{SPLATTER-BE} & AllData &   0.380 &  0.385 &  0.386 &  0.406 &      0.376 &  0.376 &  0.377 &  0.377 \\
                & PriorInfo &   0.440 &  0.423 &  0.412 &  0.405 &      0.461 &  0.473 &  0.414 &  0.398 \\
              & DTA &  {\underline{\textbf{0.387}}} (1) &  {\underline{\textbf{0.383}}} (1) &  {\underline{\textbf{0.394}}} (1) &  {\underline{\textbf{0.403}}} (1) &  {\underline{\textbf{0.383}}} (1) &  {\underline{\textbf{0.380}}} (1) &  {\underline{\textbf{0.379}}} (1) &  {\underline{\textbf{0.379}}} (1) \\
              & MA-PA &                         0.411 (3) &                         0.409 (3) &                \textbf{0.408} (2) &                \textbf{0.409} (2) &                         0.411 (3) &                         0.405 (3) &                         0.397 (3) &                         0.393 (3) \\
              & MAGAN &                         0.514 (4) &                         0.456 (4) &                         0.461 (4) &                         0.472 (4) &                         0.545 (4) &                         0.478 (4) &                         0.473 (4) &                         0.488 (4) \\
              & SSMA &                \textbf{0.410} (2) &                \textbf{0.408} (2) &                         0.408 (3) &                         0.409 (3) &                \textbf{0.393} (2) &                \textbf{0.387} (2) &                \textbf{0.386} (2) &                \textbf{0.387} (2) \\
\hline
\multirow{6}{*}{SWISSR-SCURVE} & AllData &   0.001 &  0.003 &  0.001 &  0.001 &      0.000 &  0.000 &  0.000 &  0.000 \\
& PriorInfo &   0.745 &  0.610 &  0.330 &  0.036 &      0.675 &  0.255 &  0.046 &  0.005 \\
              & DTA &                \textbf{0.097} (2) &  {\underline{\textbf{0.035}}} (1) &  {\underline{\textbf{0.004}}} (1) &                \textbf{0.002} (2) &                \textbf{0.098} (2) &  {\underline{\textbf{0.027}}} (1) &  {\underline{\textbf{0.002}}} (1) &                \textbf{0.000} (2) \\
              & MA-PA &  {\underline{\textbf{0.019}}} (1) &                         0.062 (3) &                         0.033 (3) &                         0.025 (4) &  {\underline{\textbf{0.018}}} (1) &                         0.062 (3) &                         0.037 (3) &                         0.027 (4) \\
              & MAGAN &                         0.690 (4) &                         0.564 (4) &                         0.105 (4) &  {\underline{\textbf{0.001}}} (1) &                         0.783 (4) &                         0.505 (4) &                         0.100 (4) &  {\underline{\textbf{0.000}}} (1) \\
              & SSMA &                         0.148 (3) &                \textbf{0.048} (2) &                \textbf{0.012} (2) &                         0.006 (3) &                         0.116 (3) &                \textbf{0.046} (2) &                \textbf{0.009} (2) &                         0.004 (3) \\
\hline
\end{tabular}
\end{table}

\begin{figure}[!h]
\begin{center}
\centerline{\includegraphics[width = 1.2\textwidth]{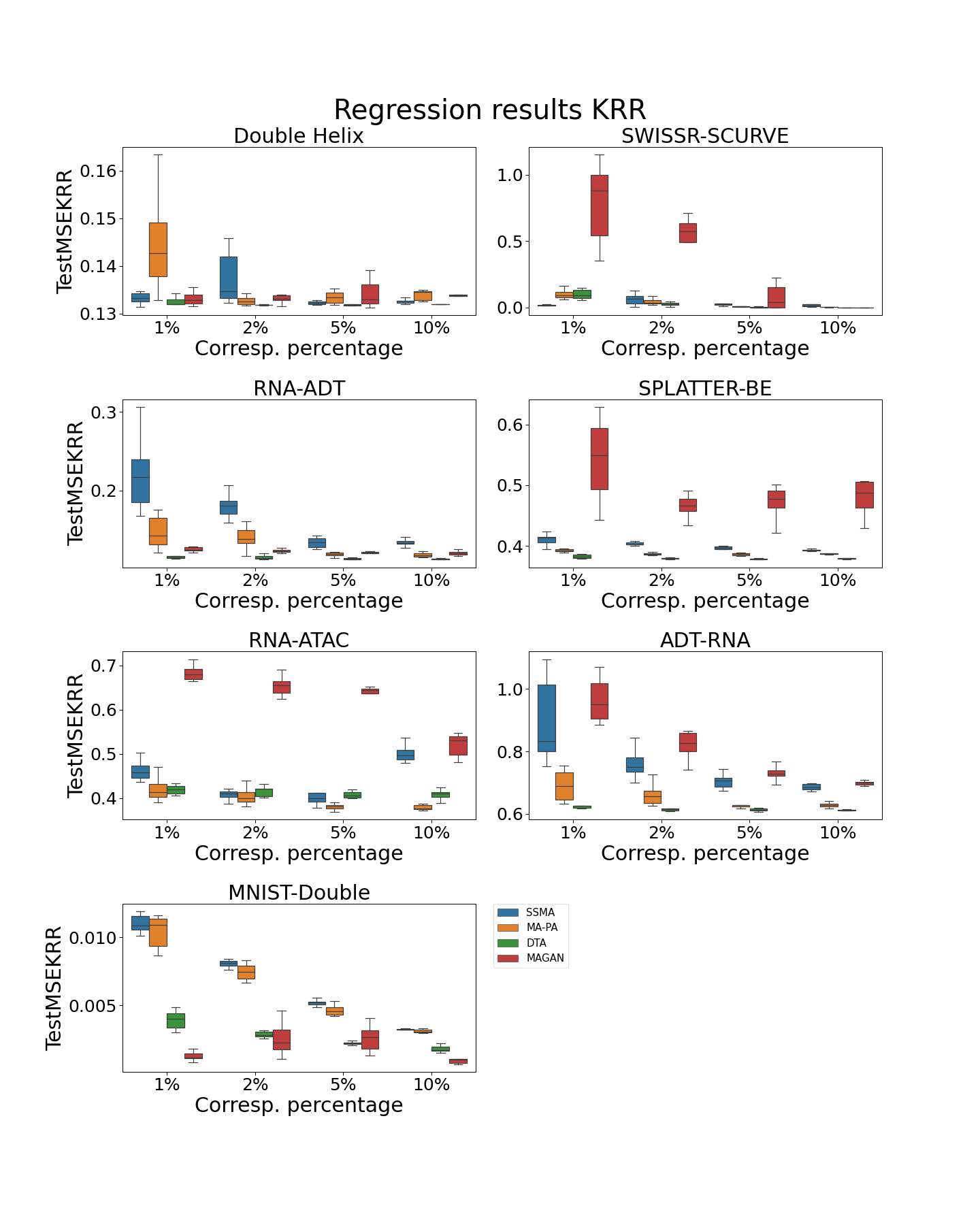}}
\caption{\textbf{Regression boxplots for the Kernel Ridge Regression (KRR) model.} The boxplots are created from 10 randomized runs for each model and for all of the datasets. See Table~\ref{tab:metrics_manifold} for the average MSE values and Figure~\ref{fig:reg_results} for a summary across datasets and across correspondence percentages.}
\label{fig:reg_boxplots2}
\end{center}
\vskip -0.2in
\end{figure}

\clearpage

\subsection{Domain adaptation results}

\begin{figure}[!h]
\begin{center}
\centerline{\includegraphics[width = 1.2\textwidth]{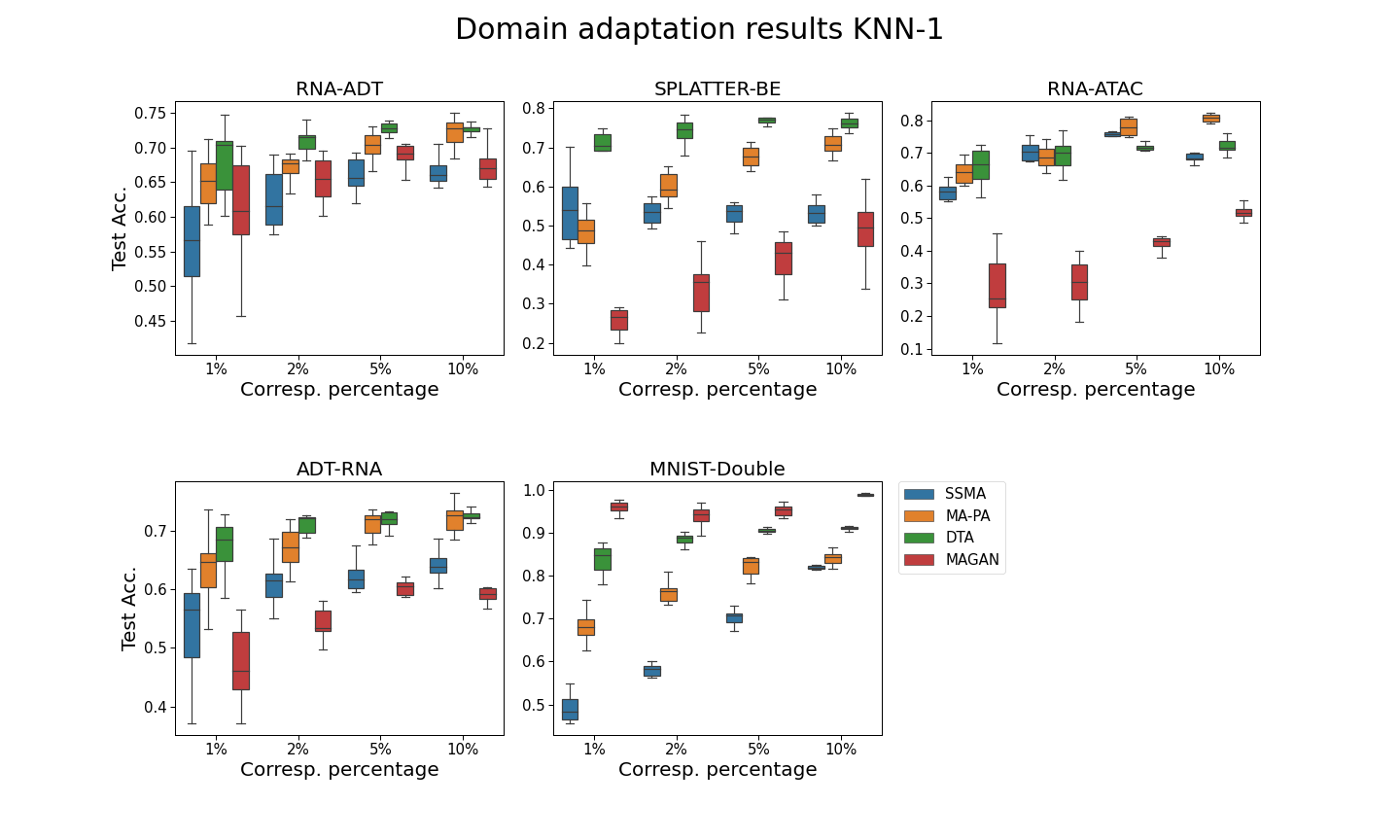}}
\caption{\textbf{Domain adaptation boxplots (KNN-1).} Test accuracy for the four manifold alignment methods, across the different datasets and various levels of known correspondences between domains. The classification is performed via a 1-nearest neighbor model trained on one domain, and then tested on the barycentric projection of the other domain on it after the alignment. DTA obtains the best results in general.}
\label{fig:class_bp1}
\end{center}
\vskip -0.2in
\end{figure}

\begin{figure}[!h]
\begin{center}
\centerline{\includegraphics[width = 1.2\textwidth]{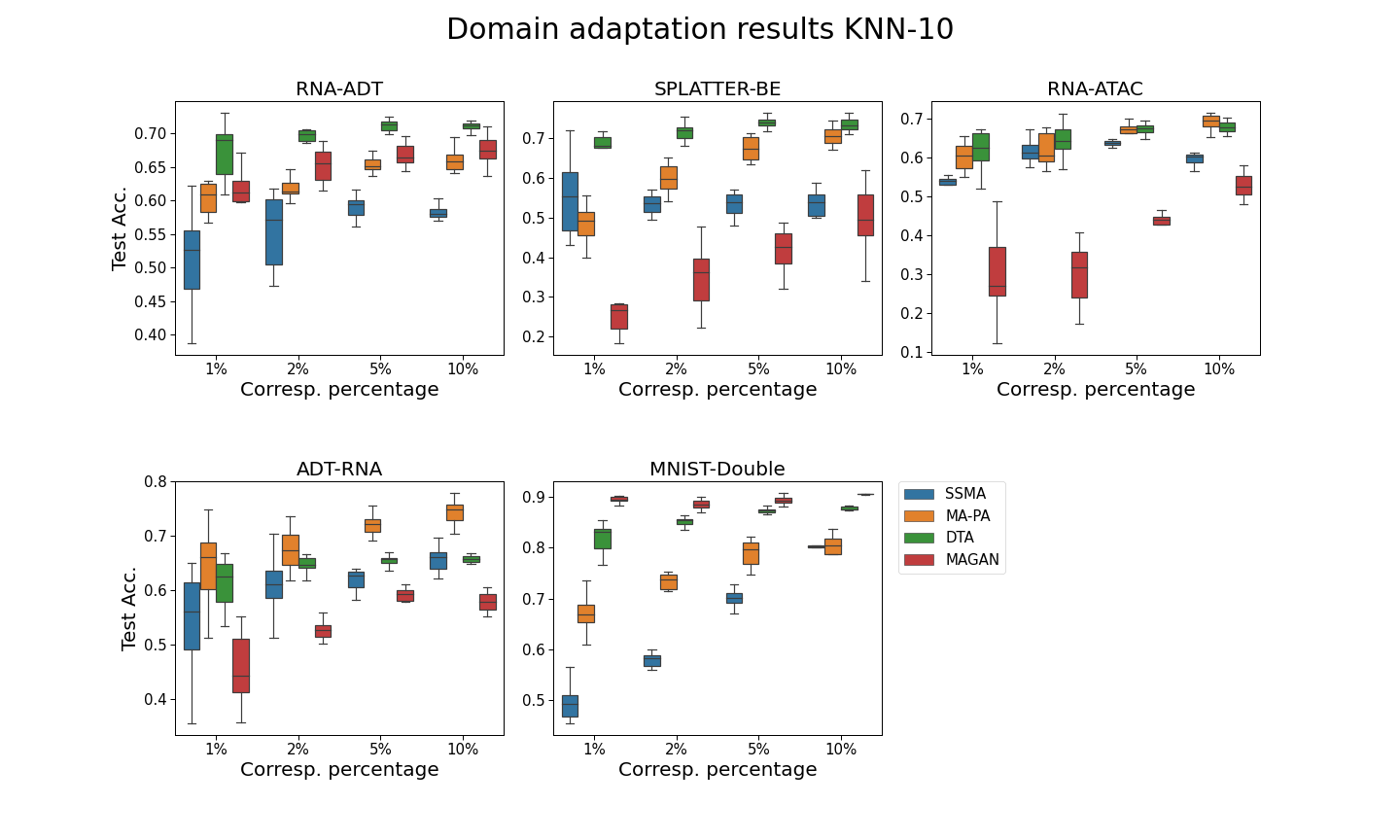}}
\caption{\textbf{Domain adaptation boxplots (KNN-10).} Boxplots for the domain adaptation problem as in Figure~\ref{fig:class_bp1}, but using a 10-nearest neighbors classifier. DTA obtains the best results in general.}
\label{fig:class_bp2}
\end{center}
\vskip -0.2in
\end{figure}

Here we provide further details of the domain adaptation results in Section~\ref{subsec:domain_adap}. Table \ref{tab:classification_results} in Section~\ref{subsec:domain_adap} shows the mean over 10 runs in terms of classification accuracy, where each run has a randomized prior known correspondences. Now we include the boxplots from those experiments in Figures \ref{fig:class_bp1} and \ref{fig:class_bp2}.  DTA obtains the best results in general as it tends to have the best mean as well as a smaller variance in the majority of scenarios.

\newpage

\subsection{Hyper-parameter selection}

Several hyper-parameters need to be selected for each model. Here we explore the hyper-parameter impact on the performance of each model for the domain adaptation experiments (see Section~\ref{subsec:domain_adap}). We focus on a correspondence percentage of 2\%.

\begin{figure}[!h]
\begin{center}
\centerline{\includegraphics[width = 1.2\textwidth]{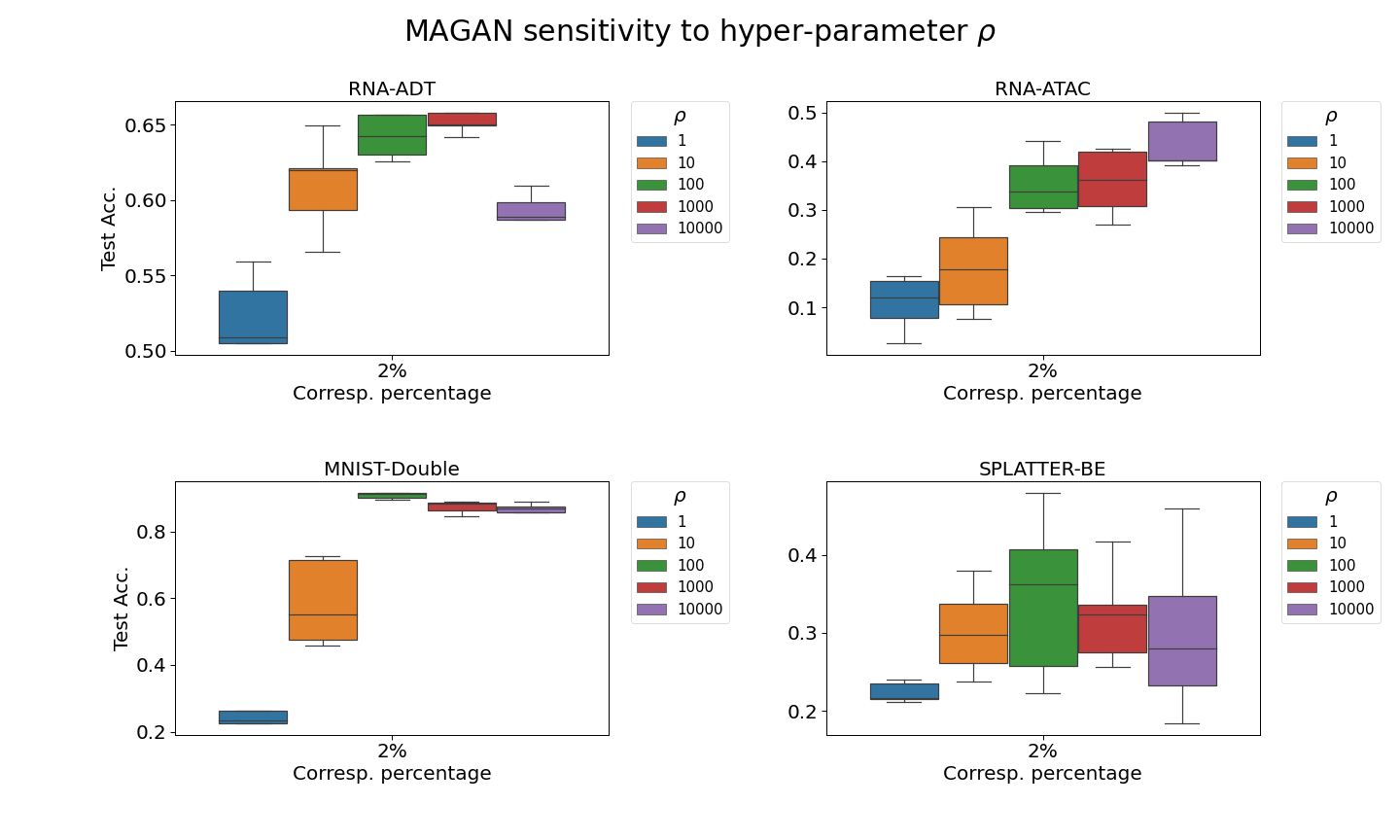}}
\caption{\textbf{Robustness test for MAGAN.} We tested the accuracy using a KNN-10 classifier for the domain adaptation problem across multiple datasets using different values of the correspondence loss regularization parameter $\rho$. For most of the datasets, the results are sensitive to the choice of $\rho$. }
\label{fig:hyp_magan}
\end{center}
\vskip -0.2in
\end{figure}

\textbf{MAGAN} is a GAN approach. Thus it requires a given architecture as well as training hyper-parameters. We originally used the code provided by the authors, but we discovered that a regularization parameter $\rho$ should be included in the correspondence loss to improve its performance. The results in Figure \ref{fig:hyp_magan} demonstrate this, where in general $\rho=100, 1000, 10000$ perform better than lower values ($\rho = 1, 10$).  Based on these results, we selected $\rho=1000$ for the other experiments in the paper.

\begin{figure}[!h]
\begin{center}
\centerline{\includegraphics[width = 1.2\textwidth]{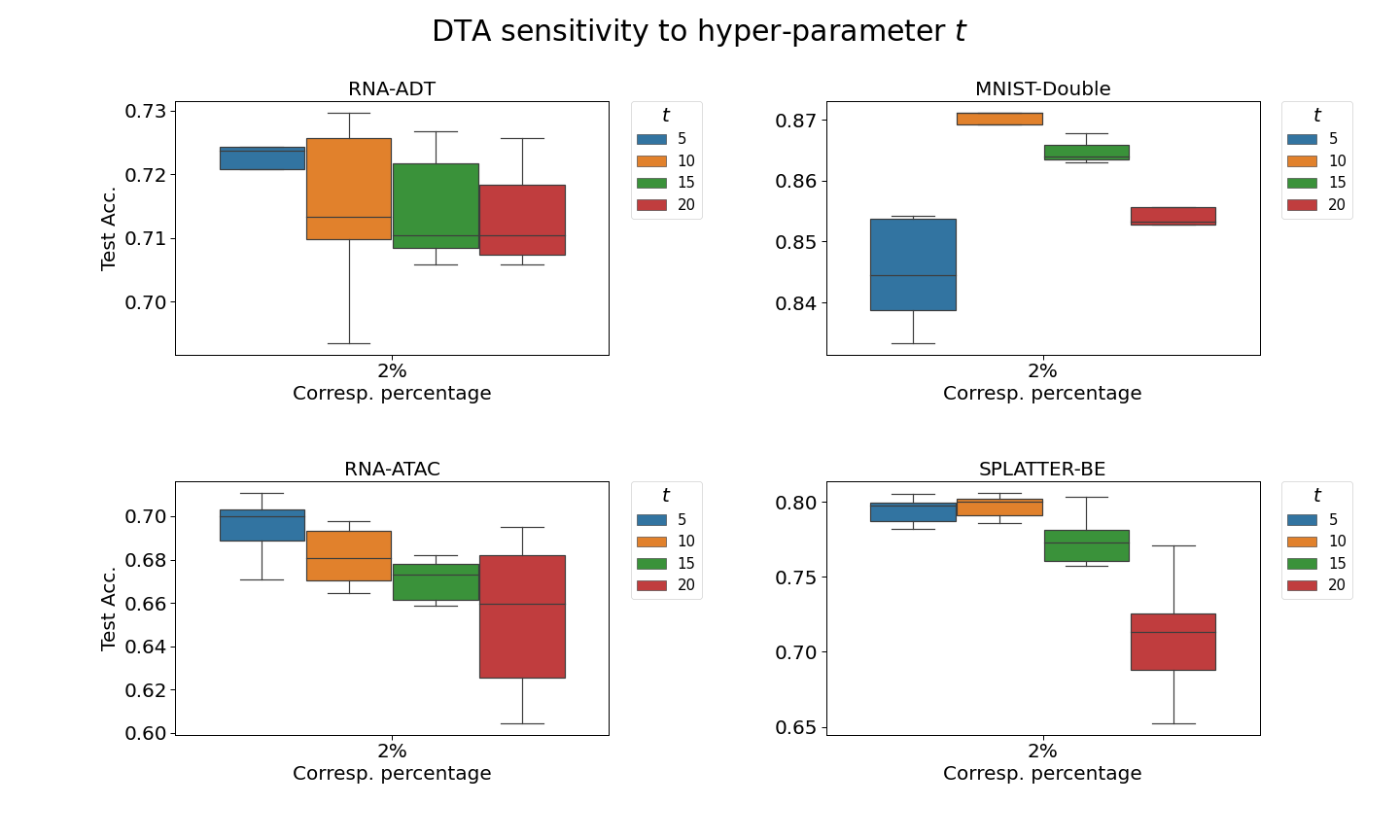}}
\caption{\textbf{Robustness test for DTA.} We tested the accuracy using a KNN-10 classifier for the domain adaptation problem across multiple datasets using different values of the diffusion time step $t$. The results are relatively robust to the choice of $t$. }
\label{fig:hyp_dta}
\end{center}
\vskip -0.2in
\end{figure}

For the graph-based methods, we fixed the $\alpha$-decay kernel with its two required hyper-parameters, $\alpha$ and $k$, as described in the main paper. Thus, we only consider one extra hyper-parameter for each of the methods as follows:

\textbf{DTA} requires one hyper-parameter $t$ that determines the number of steps in the diffusion process. Figure \ref{fig:hyp_dta} shows that DTA is relatively robust to different values of $t$, and does not suffer abrupt changes in performance. This contrasts with MAGAN which is sensitive to changes in $\rho$. Values between 5 and 15 seem to produce better results. 

\begin{figure}[!h]
\begin{center}
\centerline{\includegraphics[width = 1.2\textwidth]{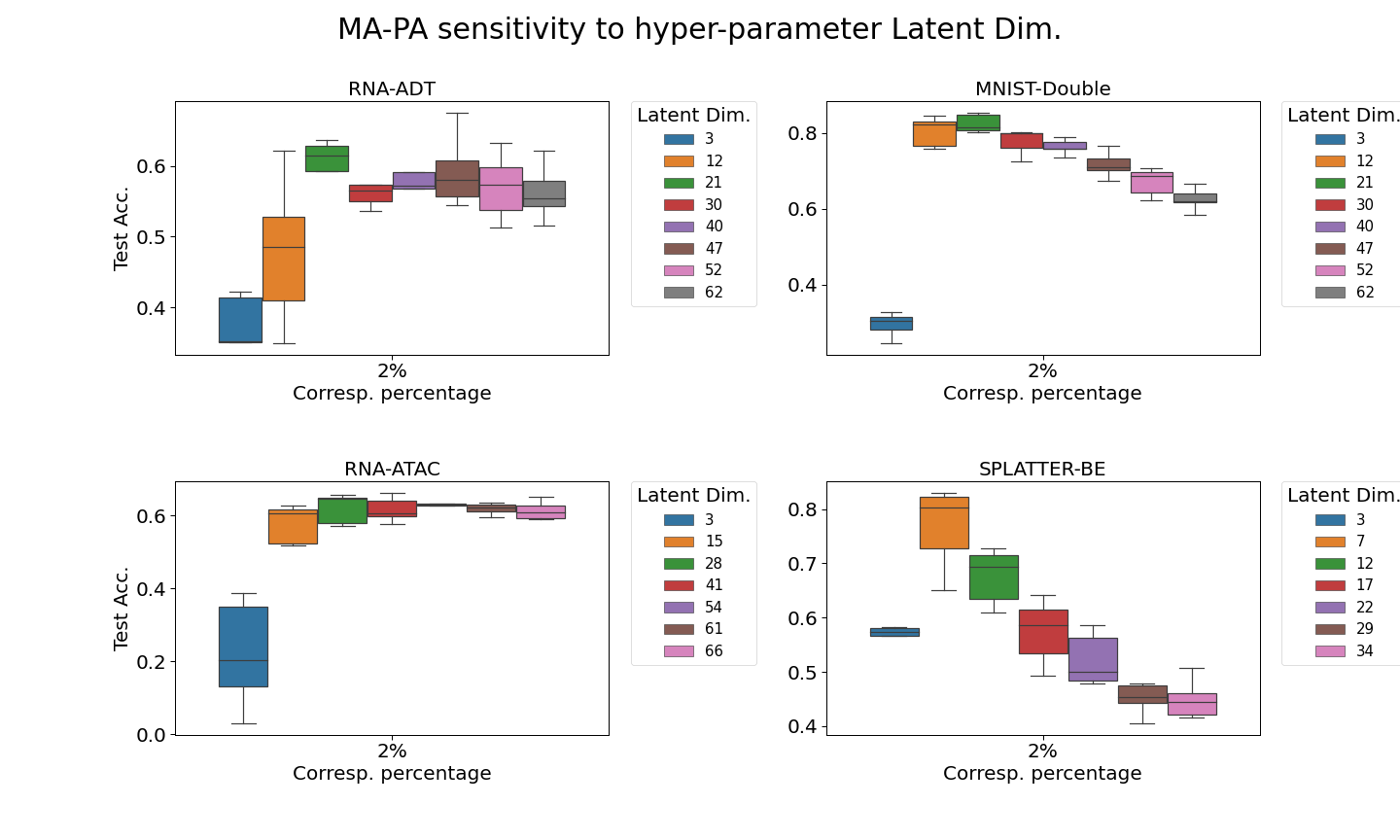}}
\caption{\textbf{Robustness test for MA-PA.} We tested the accuracy using a KNN-10 classifier for the domain adaptation problem across multiple datasets using different values of the number of latent dimensions in which the manifolds are aligned. For some of the datasets, the results  are relatively sensitive to the choice of this hyperparameter.}
\label{fig:hyp_mapa}
\end{center}
\vskip -0.2in
\end{figure}

\begin{figure}[!h]
\begin{center}
\centerline{\includegraphics[width = 1.2\textwidth]{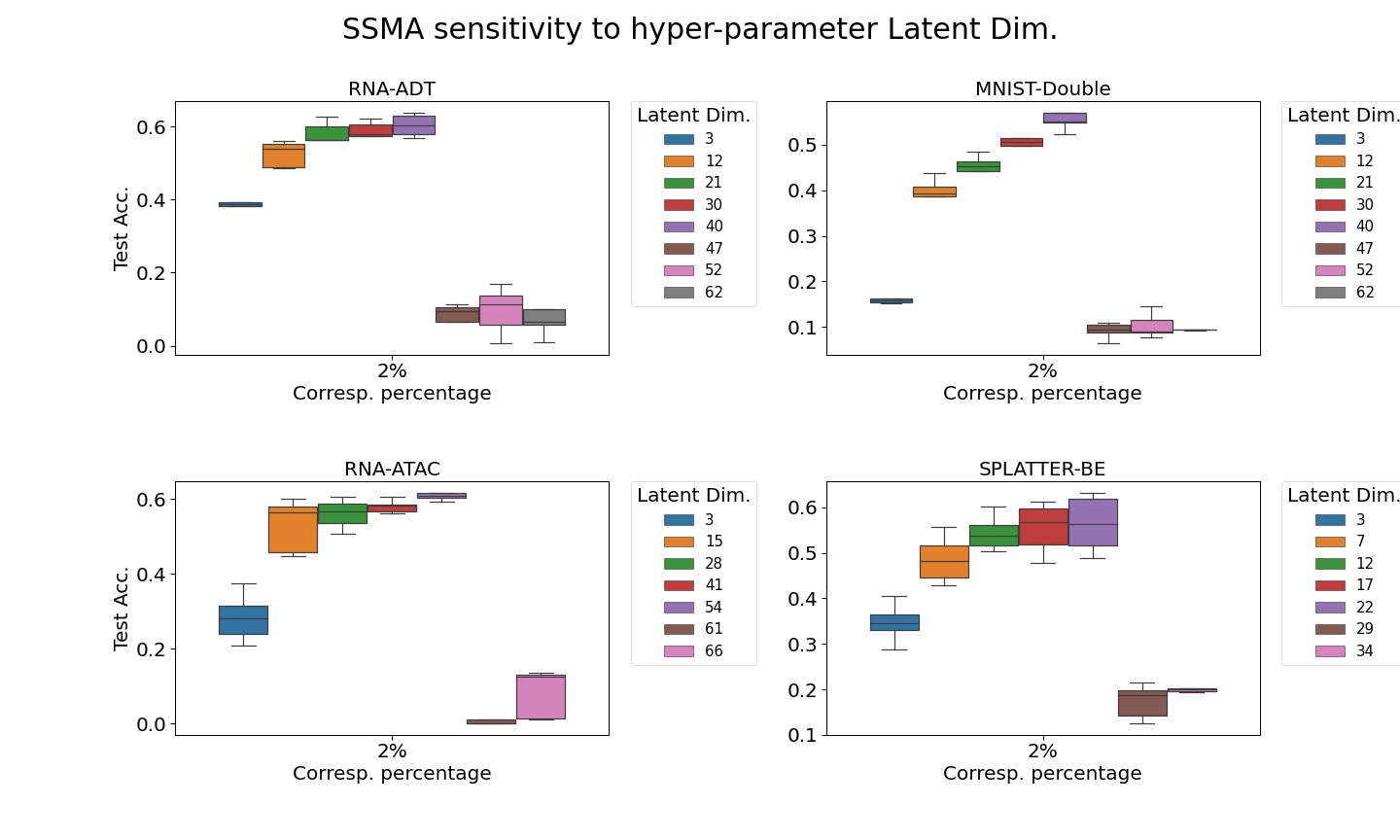}}
\caption{\textbf{Robustness test for SSMA.} We tested the accuracy using a KNN-10 classifier for the domain adaptation problem across multiple datasets using different values of  the number of latent dimensions in which the manifolds are aligned. The performance tends to increase monotonically as the number of latent dimensions increases until all of the eigenvectors associated with the nonzero eigenvalues are selected, after which there is a significant drop in performance. }
\label{fig:hyp_ssma}
\end{center}
\vskip -0.2in
\end{figure}

\textbf{MA-PA and SSMA} require  the number of latent dimensions in which the manifolds are aligned to be chosen. Their change in performance for different values is reported in Figures \ref{fig:hyp_mapa} and \ref{fig:hyp_ssma}, respectively. The behavior of SSMA is clear as the performance increases monotonically until we select all of the eigenvectors associated with the non-zero eigenvalues. In contrast, there is no consistent behavior across multiple datasets for MA-PA. We decided to use the same dimensions as that selected for SSMA, which perform decently in general except for the SPLATTER-BE dataset.

\paragraph{Computational resources:}  The experiments were performed on a personal machine equipped with 16 GB RAM and GeForce RTX 2080 Ti. According to the ML CO2 impact calculator the 100 hours of training we estimate were used in total, for the various preliminary and final experiments, correspond to approximately 10.8 kg of CO2 emitted.

\end{document}